%% file: paper.tex
\documentclass[manuscript]{acmart}
\renewcommand\footnotetextcopyrightpermission[1]{}

\usepackage[utf8]{inputenc}
\usepackage{graphicx}

\usepackage[linesnumbered, ruled, vlined]{algorithm2e}

\SetCommentSty{mycommfont}
\SetKwInput{KwInput}{Input}                
\SetKwInput{KwOutput}{Output}              

\usepackage{caption}
\usepackage{algorithmicx}
\usepackage{amsmath}
\usepackage{wrapfig}
\usepackage{amsthm}
\usepackage{mathtools}
\DeclarePairedDelimiter{\ceil}{\lceil}{\rceil}

\usepackage[export]{adjustbox}
\usepackage{xcolor}

\newif\ifsubmit
\submitfalse
\ifsubmit
    \newcommand{\todo}[1]{}
    \newcommand{\tocite}[1]{}
\else
    \newcommand{\todo}[1]{[{\color{red}TODO: #1}]}
    \newcommand{\tocite}[1]{[{\color{red}CITE: #1}]}
\fi

\newcommand{\revise}[1]{{\color{black} #1}}
\newcommand{\newrevise}[1]{{\color{black} #1}}
\newcommand{\finalrevise}[1]{{\color{black} #1}}

\title{Ax-BxP: Approximate Blocked Computation for Precision-Reconfigurable Deep Neural Network Acceleration}

\author{Reena Elangovan}
\email{elangovr@purdue.edu}
\affiliation{%
  \institution{Purdue University}
  \department{School of Electrical and Computer Engineering}
  \country{USA}
}

\author{Shubham Jain}
\email{shubham.jain35@ibm.com}
\affiliation{
  \institution{IBM T.J. Watson Research Center}
  \city{Yorktown Heights, NY}
  \country{USA}
}

\author{Anand Raghunathan}
\email{raghunathan@purdue.edu}
\affiliation{
  \institution{Purdue University}
  \department{School of Electrical and Computer Engineering}
  \country{USA}
}



\begin{document}
\begin{abstract}
\input{sections/abstract}  
\end{abstract}

\maketitle
\pagestyle{plain}

\vspace*{-5pt}
\section{Introduction}
\label{sec:introduction}
\input{sections/Introduction}


\section{Preliminaries}
\label{sec:prelims}
\input{sections/Preliminaries}

\section{Approximate Blocked Computation}
\label{sec:ApproxBlkComp}
\input{sections/blockfxp}

\section{Experimental Methodology}
\label{sec:ExpMeth}
\input{sections/exptsetup}

\section{Results}
\label{sec:Results}
\input{sections/results}

\section{Related Work }
\label{sec:RelWork}

\input{sections/relatedWork}


\vspace*{-5pt}
\section{Conclusion}
\label{sec:Conclusion}
\input{sections/conclusion}

\section*{Acknowledgement}
This work was supported by C-BRIC, one of six centers in JUMP, a Semiconductor Research Corporation (SRC) program, sponsored by DARPA.

\scriptsize
\bibliographystyle{unsrt}
\bibliography{paper}
\end{document}

%% file: sections/abstract.tex
Precision scaling has emerged as a popular technique to optimize the compute and storage requirements of Deep Neural Networks (DNNs). Efforts toward creating ultra-low-precision (sub-8-bit) DNNs for efficient inference suggest that the minimum precision required to achieve a given network-level accuracy varies considerably across networks, and even across layers within a network. This translates to a need to support variable precision computation in DNN hardware. Previous proposals for precision-reconfigurable hardware, such as bit-serial architectures, incur high overheads, significantly diminishing the benefits of lower precision. We propose Ax-BxP, a method for approximate blocked computation wherein each multiply-accumulate operation is performed block-wise (a block is a group of bits), facilitating re-configurability at the granularity of blocks. Further, approximations are introduced by only performing a subset of the required block-wise computations in order to realize precision re-configurability with high efficiency. We design a DNN accelerator that embodies approximate blocked computation and propose a method to determine a suitable approximation configuration for any given DNN. For the AlexNet, ResNet50 and MobileNetV2 DNNs, Ax-BxP achieves \revise{1.1x-1.74x and 1.02x-2x} improvement in system energy and performance respectively, over an 8-bit fixed-point (FxP8) baseline, with minimal loss (<1\% on average) in classification accuracy. Further, by varying the approximation configurations at a finer granularity across layers and data-structures within a DNN, we achieve \revise{1.12x-2.23x and 1.14x-2.34x} improvement in system energy and performance respectively. 

%% file: sections/Introduction.tex
\noindent Deep Neural Networks (DNNs) have become very popular in recent years due to their ability to achieve state-of-the-art performance in a variety of cognitive tasks such as image classification, speech recognition and natural language processing~\cite{AlphaGO,GPT3,Eff-Net}. The remarkable algorithmic performance of DNNs comes with extremely high computation and storage requirements. While these challenges span both training and inference, we focus on the latter scenario where the high computation requirements of DNN inference limit their adoption in energy- and cost-constrained devices~\cite{venkataramani2016}.

The use of low precision has emerged as a popular technique for realizing DNN inference efficiently in hardware~\cite{sze2017, TPU}. Lowering the precision or bit-width favorably impacts all components of energy consumption including processing elements, memory, and interconnect. State-of-the-art commercial DNN hardware widely supports 8-bit precision for DNN inference, and recent research continues to explore inference with even lower precision \cite{wang2019haq,mishra2017wrpn,zhou2017bq,choi2018pact,jain2019biscaled}.  

Recent efforts~\cite{wang2019haq,Proteus,PrecisionQuant} suggest that realizing ultra-low-precision (sub-8-bit) DNN inference with minimal accuracy degradation is quite challenging if the precision for all data-structures is scaled uniformly. Therefore, the use of variable precision (across DNNs, across layers within a DNN, and across different data structures) has gained considerable interest. For instance, HAQ~\cite{wang2019haq} shows that MobileNet and ResNets require precision varying from 3 to 8 bits across network layers in order to match the accuracy of a full-precision network.

\begin{wrapfigure}{r}{0.3\columnwidth}
  \includegraphics[width=0.3\columnwidth]{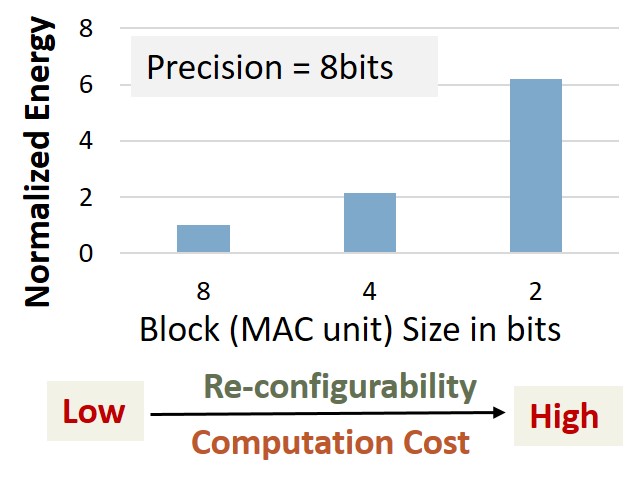}
  \caption{Computation cost vs re-configurability trade-offs}
  \label{fig:Motivation}
\end{wrapfigure}
To support variable precision, one option is to utilize conventional fixed-precision hardware that is provisioned for the highest precision, and gate the unused portions of logic to save power when computations do not use the full precision supported. However, this approach does not fully utilize the potential of aggressive precision-scaling since the unused hardware lowers utilization. Alternatively, variable-precision DNN accelerators with bit-serial~\cite{umuroglu2018bismo, judd2016stripes, loom} or bit-fused~\cite{sharma2018bit} fabrics have been designed to support re-configurability. However, this re-configurability comes at a high cost as the bit-serial arithmetic circuits incur significant energy and  latency overheads with respect to their fixed-precision counterparts of equivalent bit-width, due to multi-cycle operation and control logic~\cite{Reconf_JETACS}. 
Figure~\ref{fig:Motivation} quantifies the energy overhead (at iso-area) incurred while performing 8-bit MAC computations using digit-serial MAC units of 2-bits and 4-bits (synthesized to 15nm with Synopsys Design Compiler). As shown, the increased flexibility provided by digit-serial hardware is accompanied by a 2-6x higher energy cost when 8-bit operations must be performed. This limits the energy benefits that can be realized from variable-precision DNNs, where bit-width varies from 2-8 bits across layers and across networks~\cite{wang2019haq,Proteus,PrecisionQuant}. 

To design DNN hardware that caters to variable precision requirements with minimal overheads, we leverage the intrinsic tolerance of DNNs to approximate computation~\cite{axnn,chippa-asilomar13}, which is the basis for reduced precision itself. Specifically, we propose Ax-BxP, an approximate computation method to execute DNN inference in which weights and activations are composed of fixed-length blocks (groups of bits), and computations are performed block-wise. Approximations are introduced by only performing a subset of the block-wise computations to enable efficient re-configurability. We present a methodology to choose the best approximation configuration for each layer in a DNN. We also propose architectural enhancements to realize Ax-BxP in a standard systolic array based DNN inference accelerator with simple and low-overhead design modifications. We show that (i) Ax-BxP with varying approximation configurations across DNNs achieves \revise{1.1x-1.74x and 1.02x-2x}  improvement in system-level energy and performance respectively, and (ii) Ax-BxP with varying approximation configurations across layers within DNNs obtains \revise{1.12x-2.23x and 1.14x-2.34x} improvement in system-level energy and performance respectively, in both cases with small loss in classification accuracy (<1\% on average) with respect to an 8-bit fixed-point (FxP) baseline.

%% file: sections/Preliminaries.tex
\noindent 
\subsection{Blocked Fixed Point Representation}
\label{subsec:BxP}

The fixed-point (FxP) format is widely used for efficient realization of low-precision DNNs. Blocked fixed-point (BxP) format is an adaptation of the FxP format wherein values are partitioned into fixed-length blocks. In particular, an $(N*K)$ bit signed FxP number $X_{FxP}$ can be represented as a BxP number $X_{BxP}$ comprised of $N$ blocks of $K$ bits each, as shown in Figure \ref{fig:BxPPrelim}(a). The blocks of $X_{BxP}$ are arranged in the decreasing order of significance (place value) by default, where the significance of the $i_{th}$ block $X_i$ is $2^{i*K}$. We assume a sign-magnitude representation wherein the most significant bit within the most significant block ($X_{N-1}$) is the sign bit. The magnitude of $X_{BxP}$ denoted $|X_{BxP}|$, can be derived from its blocks as shown in the figure.


\subsection{Blocked Multiplication}
\label{subsec:blkComp}

\begin{wrapfigure}{r}{0.6\columnwidth}
 \vspace{-35pt}
 \includegraphics[width=0.6\columnwidth]{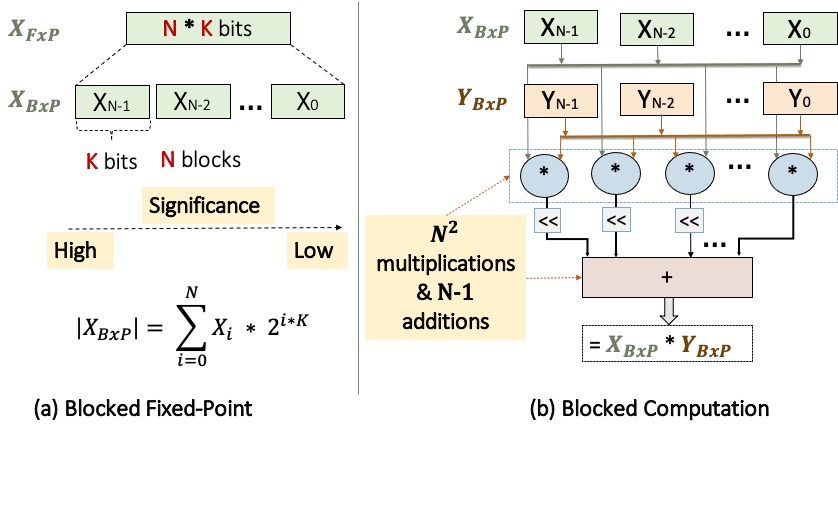}
 \vspace{-35pt}
 \caption{BxP: Overview}
 \label{fig:BxPPrelim}
\end{wrapfigure}
Figure~\ref{fig:BxPPrelim}(b) demonstrates a blocked\footnote{Blocked multiplication is also known in the literature as digit serial multiplication (e.g., \cite{DigSerial}) and bit-level composable multiplication (e.g., \cite{sharma2018bit}).} multiplication between two BxP format numbers $X_{BxP}$ and $Y_{BxP}$.  As shown, each block of $X_{BxP}$ is multiplied with each block of $Y_{BxP}$ to generate $N^2$ partial products (P). Subsequently, the partial products are shifted and accumulated using $N^2-1$ additions. Equation~\ref{eq:BxP_mult} expresses an exact blocked multiplication operation, where $P_{ij}$ [$P_{ij}$ = $X_{i}$ * $Y_{j}$] represents the partial product of the $i^{th}$ block of $X_{BxP}$ ($X_{i}$) and $j^{th}$ block of $Y_{BxP}$ ($Y_{j}$).

\begin{equation} 
\centering
\label{eq:BxP_mult}
\begin{aligned}
X_{BxP} * Y_{BxP} = \sum_{i=0}^{i=N-1}\sum_{j=0}^{j=N-1}  P_{ij}*2^{(i+j)*K}\\
\end{aligned}
\end{equation}

%% file: sections/blockfxp.tex
In this section, we discuss Ax-BxP, the proposed approximate blocked computation method for designing efficient precision-reconfigurable DNNs. We first detail the key approximation concepts and introduce the Ax-BxP format for representing MAC operands. Subsequently, we present a systematic methodology for designing Ax-BxP DNNs with minimal impact on application-level accuracy. Finally, we demonstrate the integration of Ax-BxP into a standard systolic array-based DNN accelerator using simple hardware enhancements.

\begin{figure}[htb]
  \vspace*{-0pt}
  \centering
  \includegraphics[width=\columnwidth]{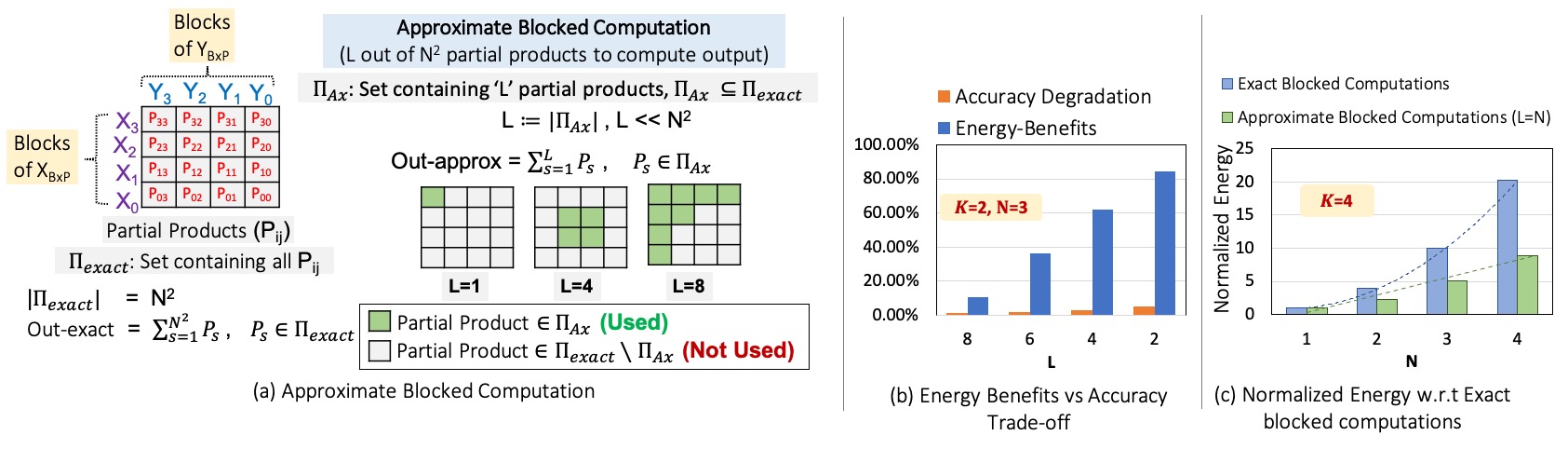}
  \vspace*{-4pt}
  \caption{Approximate blocked computation: Overview}
  \label{fig:Overview}
  \vspace*{-4pt}
\end{figure}
\subsection{Approximation Method}
\label{subsec:ApproxMeth}

The main idea in Ax-BxP is to perform blocked computation by performing only a subset of the required block-wise computations. Figure~\ref{fig:Overview}(a) illustrates the concept, where the multiplication of operands ($X_{BxP}$ and $Y_{BxP}$) is performed by computing and accumulating only $L = |\Pi_{Ax}|$ out of the total of $N^2$ partial product terms. Formally, we characterize Ax-BxP multiplication using a set $\Pi_{Ax} \subseteq \Pi_{Exact}$, wherein the final output (out-approx) is given by summing the partial products ($P_{s}$) in $\Pi_{Ax}$. Ax-BxP involves two key design choices --- (i) $L$ (the size of set $\Pi_{Ax}$) and (ii) the choice of partial products to include within the set $\Pi_{Ax}$. These design choices affect both the computational errors introduced and the energy benefits, and therefore need to be explored judiciously to produce the best possible energy-accuracy trade-offs. 

Figure \ref{fig:Overview}(b) presents the energy-accuracy tradeoff provided by Ax-BxP for the AlexNet DNN on the CIFAR10 dataset, across various choices of $L$ with fixed values of N and K (note that $K=2, N=3$, with $L=9$ corresponds to exact blocked computation). As $L$ decreases, the accuracy decreases minimally, whereas the energy benefits increase drastically. The computational accuracy reported is for the best choice of $\Pi_{Ax}$ among all choices of $\Pi_{Ax}$, identified using the methodology described in Section~\ref{subsec:DesMeth}. To estimate these energy benefits, we synthesized the exact and approximate RTL designs (described in Section~\ref{subsec:accdes}) to the 15nm technology node using Synopsys Design Compiler. The results suggest a favorable energy-accuracy trade-off, which arises due to the typical data distribution seen in DNNs~\cite{jain2019biscaled}, wherein  a majority ($>$90\%) of the operations can be computed accurately at low-precision, but computing the remaining operations at higher precision is critical to preserving accuracy. We also evaluated the energy benefit across various values of N by fixing the values of K and L (K=4, L=N). As shown in Figure~\ref{fig:Overview}(c), the energy of approximate block-wise computations increase linearly with N as we require O(N) block-wise partial products to be evaluated and accumulated. In contrast, the energy of exact blocked computation increases quadratically as it requires $O(N^2)$ block-wise partial products. 

For a given $L$, when fewer than $N$ blocks of $X_{BxP}$ or $Y_{BxP}$ (or both) are used to construct $\Pi_{Ax}$, we can achieve memory footprint savings in addition to computation savings by storing only the required blocks. To leverage this, we introduce a new Ax-BxP tensor format for storing the Ax-BxP operands in the following sub-section. Section \ref{subsec:DesMeth} describes the \textit{significance-} and \textit{value-based} methods that we use for choosing the required blocks of the operands.

\subsection{Ax-BxP Tensor}
\label{subsec:AxBxP}
\begin{wrapfigure}{r}{0.4\columnwidth}
 \vspace*{-20pt}
 \centering
 \includegraphics[width=0.4\columnwidth]{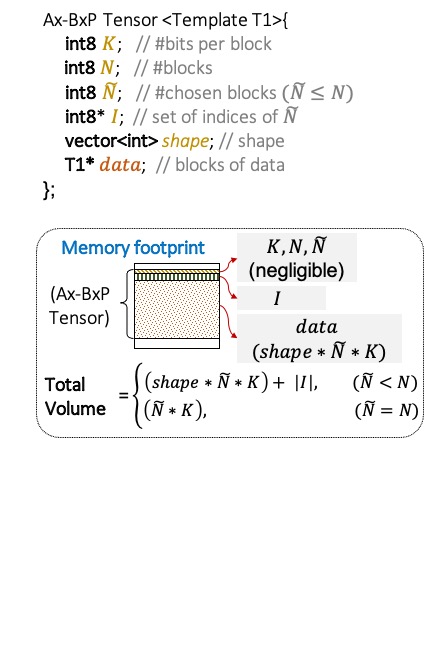}
 \vspace*{-100pt}
 \caption{Ax-BxP Tensor: Memory Layout}
 \label{fig:AxBxPtensor}
 \vspace*{-0pt}
\end{wrapfigure}
We introduce a new format for storing Ax-BxP operand blocks and their indices. The Ax-BxP format uses the following fields -- (i) total number of blocks ($N$), (ii) block-size ($K$), (iii) the set $\mathcal{I}$ containing indices of the operand blocks chosen for Ax-BxP computation, (iv) number of chosen blocks ($\tilde{N} \coloneqq |\mathcal{I}| \leq N$) and (v) data blocks (\emph{data}), which are arranged in decreasing order of significance. The size of the \emph{data} field is  $\tilde{N}*K$ bits. Note that, during exact blocked computation, $\tilde{N} = N$ and $\mathcal{I}$ is not required. 

An Ax-BxP Tensor (typically, the weights or activations of an entire layer) is composed of scalar elements in the Ax-BxP Format, where the elements share common values of the parameters (\emph{viz.}, N, K, $\tilde{N}$, $\mathcal{I}$). Ax-BxP Tensor is presented as a C++ template class in Figure \ref{fig:AxBxPtensor}, along with an illustrative memory layout of the different fields. Since the space required to store the parameters is amortized across an entire tensor, the associated memory footprint is negligible. Furthermore, when $\tilde{N}<N$ the size of the \emph{data} field is reduced, resulting in savings in memory footprint and memory traffic in addition to computation.

\subsection{Design Methodology}
\label{subsec:DesMeth}

In this sub-section, we present the key design considerations involved in approximating DNNs using Ax-BxP and the methodology we use to select the Ax-BxP format for each layer in the network. We first characterize the Ax-BxP design space. Subsequently, we provide pruning techniques to reduce the complexity of exploring this design space and algorithms for the systematic design of Ax-BxP DNNs.


\subsubsection{Design Space}
\label{subsubsec:DesSpaceChar}

For a given bit-width (BW), where BW $=$ N * K, an Ax-BxP MAC operation (characterized by the set $\Pi_{Ax}$) can be designed in numerous ways. We define $\Omega$ as a set enumerating all possible ways of constructing $\Pi_{Ax}$. Equation~\ref{eq:DesignSpace} expresses the size of $\Omega$ ($|\Omega|$) which is determined by free variables L and N. As shown, for a given BW, we are free to choose N (i.e., number of blocks) to be an integer from 1 to BW. Subsequently, we can select the approximation-level by determining L, i.e., number of partial products to be used during MAC operations. The value of L can be 1 to $N^2$, where L=$N^2$ represents an exact blocked computation. Lastly, there are $N^2 \choose L$ ways of selecting L out of  $N^2$ partial products.

\begin{equation} 
\centering
\label{eq:DesignSpace}
\begin{aligned}
    |\Omega| = \sum_{N=1}^{BW}\sum_{L=1}^{N^2} {N^2 \choose L} \\
\end{aligned}
\end{equation}

\subsubsection{Design Space Constraints}
\label{subsubsec:DSC}

To reduce the search space $|\Omega|$ which is exponential in N, we put forth the following arguments to bound $|\Omega|$ by constraining BW, K, N, and L:

\begin{itemize}
    \item \emph {Bitwidth (BW)}: Since a bit-width of 8 for both activations and weights is sufficient to preserve accuracy during inference~\cite{zhou2016dorefanet}, we constraint BW $\leq$ 8. 
    \item \emph{Bits in a block ($K$)}: We also bound $K$ such that $1 < K \leq 4$. By setting $K > 1$, we avoid the latency and energy overheads associated with bit-serial (K=1) implementations~\cite{sharma2018bit}. Moreover, we introduce an upper-bound on $K$ ($K \leq 4$) to avoid $N = 1$ (i.e., an FxP implementation).
    \item \emph{Number of Blocks ($N$)}: We set N=$\ceil{BW/K}$. Therefore, for BW $\leq$ 8 and $1 < K \leq 4$, the allowed values of N are 2,3 and 4. 
    
     \item \emph{Size of set $\Pi_{Ax}$ (L)}: Lastly, we constraint L $\leq$ N based on the energy-accuracy trade-offs discussed in Section~\ref{subsec:ApproxMeth} and shown in Figure~\ref{fig:Overview}. We found that L $\leq$ N provides ample design choices, wherein we can obtain significant energy benefits with minimal impact on computational accuracy. Apart from a reduction in design space, bounding L $\leq$ N also helps in minimizing the design complexity of both the control logic and Ax-BxP PEs at the systolic-array level, mitigating the associated reconfiguration overheads (discussed further in Section~\ref{subsec:accdes}). 
\end{itemize}

Equation~\ref{eq:DesignSpaceConst} expresses the size of reduced design space ($\Omega_{c}$) obtained after constraining the variables BW, K, N, and L. 
\begin{equation} 
\centering
\label{eq:DesignSpaceConst}
\begin{aligned}
|\Omega_{c}| = \sum_{N=2,3,4}\sum_{L=1}^{N} {N^2 \choose L} 
\end{aligned}
\end{equation}
Using numerical methods, we evaluate $|\Omega_{c}|$ to be \revise{$2655$} for a single DNN layer. For a DNN with $n$ layers, $|\Omega_{c}|$ = \revise{$2655^n$}. This space is too large to be exhaustively explored for practical DNNs, especially since re-training is needed to alleviate the accuracy degradation caused by approximation. We therefore propose techniques to prune the search space by eliminating sub-optimal configurations.

\subsubsection{Design Space Pruning}
\label{subsubsec:Prun}

\begin{figure}[htb]
 \vspace*{-0pt}
 \centering
 \includegraphics[width=0.9\columnwidth]{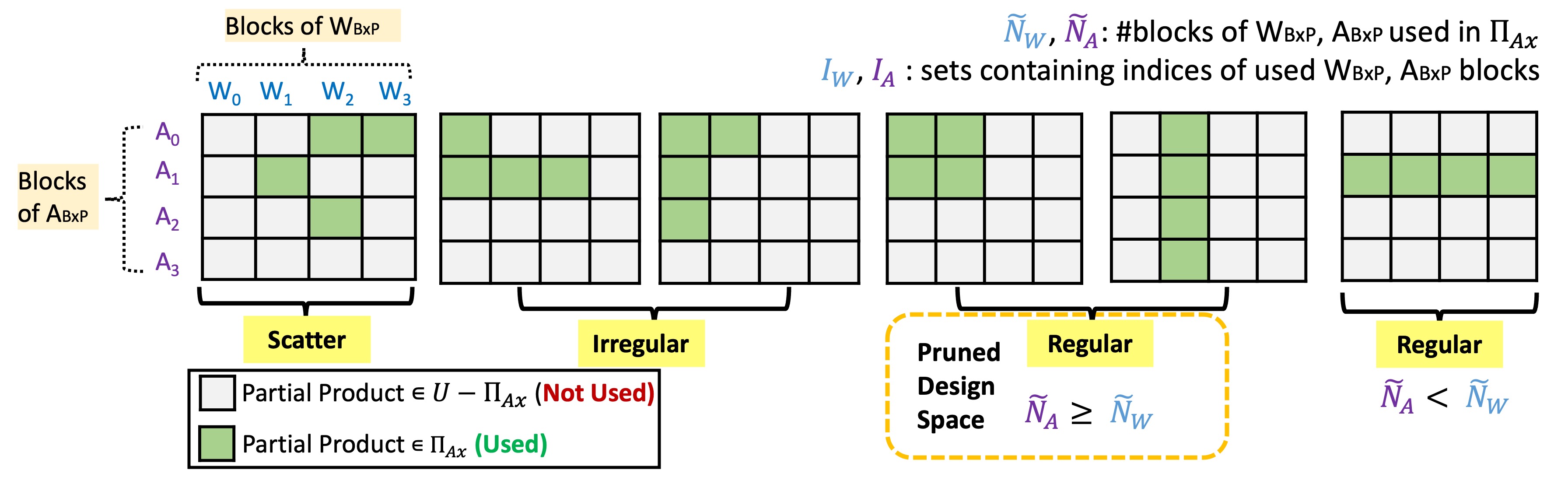}
 \vspace*{-0pt}
 \caption{Design search space for L=4, N=4: Illustration}
 \label{fig:blockShape}
 \vspace*{-0pt}
\end{figure}

We prune the search space by restricting the contents of the set $\Pi_{Ax}$ for a given L. Figure~\ref{fig:blockShape} illustrates the possible choices of $\Pi_{Ax}$, wherein $W_{BxP}$ and $A_{BxP}$ are the Ax-BxP weight and activation tensors, respectively, of a DNN layer. Further, $\tilde{N}_W$ ($\tilde{N}_A$) and  $\mathcal{I}_W$ ($\mathcal{I}_A$) represent number of blocks and block indices respectively, of $W_{BxP}$ ($A_{BxP}$) used for computing the MAC operation ($W_{BxP}$*$A_{BxP}$). As shown, there are a variety of ways of choosing 4 (L) out of 16 ($N^2$) partial products. The selected partial products (shown in green) cast a shape on the 2D array that represents all possible $N^2$ partial products. The shape could be scattered, irregular, or regular. In our design exploration, we restrict ourselves to regular shapes, which substantially reduces the search space complexity. Formally, restricting to regular shapes constraints L as shown in Equation~\ref{eq:lcons1}. 

\begin{equation} 
\centering
\label{eq:lcons1}
\begin{aligned}
L = \tilde{N}_W * \tilde{N}_A. 
\end{aligned}
\end{equation}
 
Based on previous studies that show activations to be more sensitive to precision-scaling than weights during DNN inference operations~\cite{zhou2016dorefanet, choi2018pact}, we further prune the search space such that $\tilde{N}_A \geq \tilde{N}_W$. In other words, we never select configurations (e.g., the right most configuration in Figure~\ref{fig:blockShape}), wherein a weight operand ($W_{BxP}$) has more blocks than an activation operand ($A_{BxP}$). Equation~\ref{eq:DesignSpacePrun} shows the size of the design search space ($\Omega_{c+p}$) obtained after pruning, wherein $N \choose \tilde{N}_W$ and $N \choose \tilde{N}_A$ are the number of possible ways of selecting $\tilde{N}_W$ and $\tilde{N}_A$ blocks, respectively, out of N blocks. It is worth mentioning that the use of regular configurations enables an efficient systolic array implementation so that we can accrue benefits from Ax-BxP due to reduced computation, memory footprint and memory traffic. 

\begin{equation} 
\centering
\label{eq:DesignSpacePrun}
\begin{aligned}
|\Omega_{c+p}| = \revise{\sum_{N=2,3,4}\sum_{\tilde{N}_A=1}^{N}\sum_{\tilde{N}_W=1}^{\tilde{N}_A} {N \choose \tilde{N}_A}{N \choose \tilde{N}_W}}
\end{aligned}
\end{equation}

\subsubsection{Design heuristics}
\label{subsubsec:designheuristics}

Next, we present the two heuristics, viz., \emph{static-idx} and \emph{dynamic-idx}, to select $\tilde{N}_W$ ($\tilde{N}_A$) blocks of $W_{BxP}$ ($A_{BxP}$).

In the \emph{static-idx} heuristic, the operand blocks are chosen in a \emph{significance-aware} manner where the blocks of higher significance are always chosen over the blocks of lower significance. For a given $\tilde{N}$, we first find the index of the most-significant non-zero block of the operand tensor and choose the next $\tilde{N}$ consecutive blocks in the decreasing order of significance. Since the blocks of data in the \emph{data} field of the Ax-BxP tensor are arranged in the decreasing order of significance by default, we require only the start index ($\mathcal{I}[N-1]$) or the end index ($\mathcal{I}[0]$) to determine the indices of all the blocks. Recall that $\mathcal{I}$ is common to all the scalar elements of an operand tensor. 

\begin{wrapfigure}{r}{0.4\columnwidth}
\includegraphics[width=0.4\columnwidth]{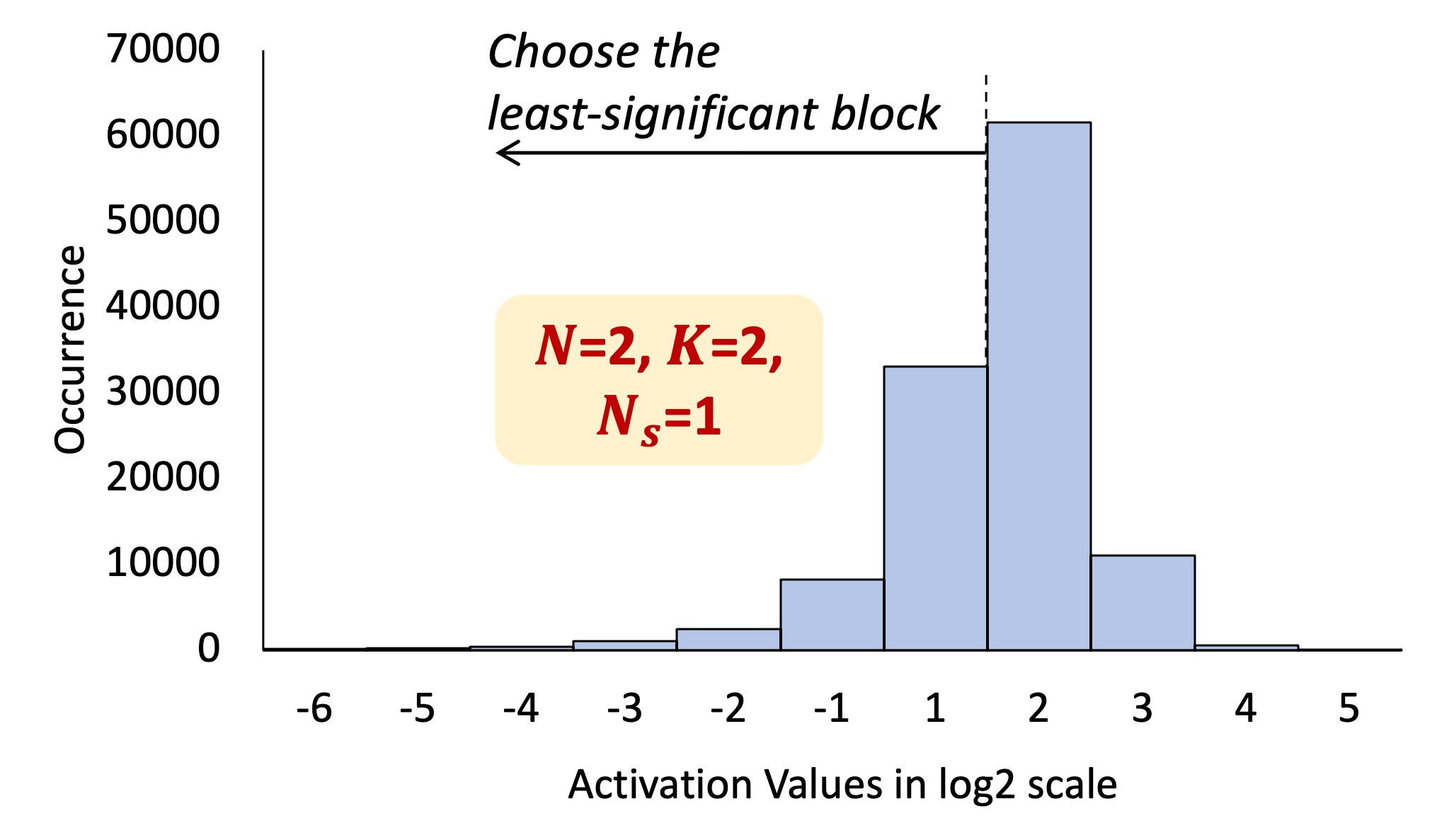}
 \caption{Significance of the non-zero activation blocks in a layer of quantized AlexNet}
 \label{fig:Act_Stat}
 \vspace{-10pt}
\end{wrapfigure}
Note that when $\mathcal{I}$ is chosen using the \emph{static-idx} heuristic, the small-valued scalar elements in an operand tensor cannot be represented with sufficient resolution. For instance, consider the activation histogram (generated using the ImageNet dataset) of a quantized AlexNet layer shown in Figure \ref{fig:Act_Stat}, where the activation values are represented using $2$ blocks of $2$ bits each, (i.e $N=2, K=2$). In the histogram, both blocks may be non-zero for large values that reside in bins [2,4], whereas only the least significant block is non-zero for the values in bins [-3,1]. Therefore, when $\tilde{N}=1$, the \emph{static-idx} heuristic chooses the most significant block for each scalar element, and the small values in bins [-3,1] are approximated to zero. To avoid this loss of information in the smaller-value scalar elements, we introduce the \emph{dynamic-idx} heuristic, where the set $\mathcal{I}$ is chosen specifically for each scalar element in a tensor.

\begin{wrapfigure}{r}{0.4\columnwidth}
 \includegraphics[clip,width=0.4\columnwidth]{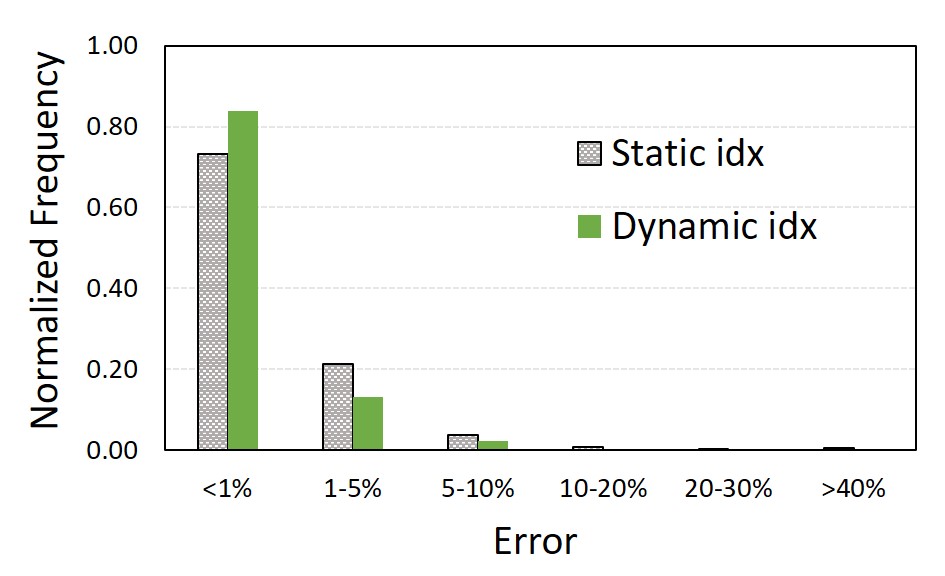}
 \caption{Operand blocks selection heuristics: Error Analysis}
 \label{fig:errorAnalysis}
\end{wrapfigure}
In the \emph{dynamic-idx} heuristic, $\mathcal{I}$ (i.e., the index of the most significant non-zero block) is determined for each scalar value of a tensor. Subsequently, each scalar is represented using $\tilde{N}$ consecutive blocks starting from it's most-significant non-zero block. Note that the dynamic heuristic is based on both significance and value of the blocks. Figure \ref{fig:errorAnalysis} quantifies the advantage of the \emph{dynamic-idx} heuristic over the \emph{static-idx} heuristic by showing the error distribution in the activations (obtained using the ImageNet dataset) for an Alexnet layer. As shown, \emph{dynamic-idx} can achieve noticeably lower error in comparison to \emph{static-idx}. On the other hand, memory savings are less for the \emph{dynamic-idx} heuristic in comparison to the \emph{static-idx} heuristic, as we need to store $I$ for each scalar element. For a given $N$ and $\tilde{N}$, the overhead of appending $\mathcal{I}$ to each data element is $\ceil{log2((N-\tilde{N}) + 1)}$ bits. For example, if K=2, N=4, $\tilde{N}$=2, we have 2 additional index bits for every 4 (K*$\tilde{N}$) compute bits. Therefore, the overall memory footprint decreases by $\sim$25\% (from 8 bits to 6 bits) compared to 8-bit fixed point. On the other hand, with \emph{static-idx}, the same configuration would require only 4 bits, resulting in a $\sim$50\% memory savings.

Since $L$ and the set $\Pi_{Ax}$ can be derived for a given $K$, $\tilde{N}_W$ and $\tilde{N}_A$, the design space $\Omega$ can now be restricted to the set of all 3-tuples \{$K$, $\tilde{N}_W$, $\tilde{N}_A$\} that satisfy all the constraints discussed thus far. The Ax-BxP configurations in $\Omega$ are listed against the block-size $K$ in Table \ref{tab:DesSpace}. Furthermore, we define two modes of Ax-BxP \emph{Static} and \emph{Dynamic} where the operand blocks are chosen using the \emph{static-idx} heuristic and the \emph{dynamic-idx} heuristic respectively.

\begin{table}[htb]
    \centering
    \caption{Design Space for approximate blocked computation}
    \begin{tabular}{|c|c|}
    \hline
         Block-size & Ax-BxP configuration = \{($K$, $\tilde{N}_W$, $\tilde{N}_A$)\} \\
         \hline
         $K=2$ & \{(2,1,4), (2,1,3), (2,2,2), (2,1,2), (2,1,1)\} \\ 
         $K=3$ & \{(3,1,3), (3,1,2), (3,1,1)\} \\ 
         $K=4$ & \{(4,1,2), (4,1,1)\} \\
         \hline
    \end{tabular}
    \label{tab:DesSpace}
    \vspace{-8pt}
\end{table}

\vspace{4pt}
\subsubsection{Designing DNNs using Ax-BxP}
\label{subsubsec:algorithms}

We now present a systematic methodology to design DNNs using Ax-BxP. Algorithm~\ref{algo:Des_Ax-BxP_DNN} describes the pseudo code that we utilize to identify best Ax-BxP configuration for each data-structure of each DNN layer. It takes a pre-trained DNN model ($DNN_{FxP}$), a training dataset ($Tr_{data}$), a target block size ($K_{tgt}$), and a limit on allowed accuracy degradation ($\gamma$) as inputs and produces an Ax-BxP DNN ($DNN_{AxBxP}$) as output. We first utilize $Tr_{data}$ to evaluate the baseline network accuracy \revise{(line 4)} and construct data-value histograms ($DsHist_{list}$) of each data-structure within the network \revise{(line 5)}. Next, we identify the best Ax-BxP configuration for a DNN layer using the histograms of the associated weight ($W_{FxP}$) and activation ($A_{FxP}$) pair \revise{(lines 6-12)}. As detailed in Algorithm~\ref{algo:Des_Ax-BxP_DNN}, to obtain best Ax-BxP configuration, we first form a pruned search space ($\Omega_{c+p}$) \revise{(line 7)}, and subsequently, explore the choices within $\Omega_{c+p}$ to find the best Ax-BxP configuration which is represented by $N_{W}$, $\tilde{N}_W$, $N_{A}$, and $\tilde{N}_A$ \revise{(line 8)}. Next, data-structures are converted to Ax-BxP tensor using the Convert-To-AxBxP function \revise{(lines 9-10)} and inserted into $DNN_{AxBxP}$ network \revise{(line 11)}. Once all data-structures are converted to Ax-BxP tensor, we re-train $DNN_{AxBxP}$ until the network accuracy is within the desired degradation limit ($<\gamma$) or the maximum allowed trained epochs (maxEpoch) is exceeded.

\begin{algorithm}
\caption{Designing AxBxP DNN}
\label{algo:Des_Ax-BxP_DNN}
\begin{algorithmic}[1]
\State \textbf{Input:}\{$DNN_{FxP}$: Pre-trained FxP DNN, $Tr_{data}$: Training dataset, $\gamma$: Max Accuracy Loss, $K_{tgt}$: Target block size\}
\State \textbf{OUTPUT:} Approximate Blocked DNN

\State {$DNN_{AxBxP}$ = $DNN_{FxP}$ } \tcc{initialize to the given FxP DNN}
\State { $Acc_{FxP}$ = computeAccuracy($DNN_{FxP}$)}
\State {$DsHist_{list}$ = getDist($DNN_{FxP}$, $Tr_{data}$) $\forall$ datastructures}
\State {{\bf for each} [Weight ($W_{FxP}$), Activations ($_{FxP}$)] pair $\in$ $DsHist_{list}$}
\State {\quad $\Omega_{c+p}$ = formPrunedSearchSpace ($W_{FxP}$, $A_{FxP}$, $K_{tgt}$)}
\tcc{Form the pruned search space of Ax-BxP configurations}
\State {\quad ($N_{W}$, $\tilde{N}_W$, $N_{A}$, $\tilde{N}_A$) = getBestConfig ($\Omega_{c+p}$, $DNN_{AxBxP}$, $K_{tgt}$, $\gamma$)} \tcc{Search for the best Ax-BxP configuration}
\State {\quad $W_{AxBxP}$ = Convert-To-AxBxP ($W_{FxP}$,$N_{W}$,$\tilde{N}_W$, $K_{tgt}$)}
\State {\quad $A_{AxBxP}$ = Convert-To-AxBxP ($A_{FxP}$, $N_{A}$, $\tilde{N}_A$, $K_{tgt}$)}
\State {\quad insert-AxBxP-Tensors ($DNN_{AxBxP}$,$W_{AxBxP}$,$A_{AxBxP}$)}
\State {{\bf end for}}
\State {numEpochs=0}
\State {{\bf while} $\gamma$ $<$ ($Acc_{FxP}$ - computeAccuracy($DNN_{AxBxP}$)   {\bf and} numEpochs $<$ maxEpoch  }) \tcc{Re-train to recover accuracy}
\State {\quad AxBxP-Aware-Training ($DNN_{AxBxP}$, $Tr_{data}$)}
\State {\quad numEpochs++}
\State {return $DNN_{AxBxP}$}
\end{algorithmic}
\end{algorithm} 

\begin{algorithm}
\revise{
\caption{getBestConfig}
\label{algo:getBestConfig}
\begin{algorithmic}[1]
\State \textbf{Input:} \{$\Omega_{c+p}$: Pruned Design Space, $DNN_{AxBxP}$, $K_{tgt}$: Target block size, $\gamma$: Max Accuracy Loss\}
\State \textbf{OUTPUT:} Best AxBxP Config
\State {{\bf while} $\gamma$ $<$ ($Acc_{FxP}$ - $Acc_{AxBxP}$)}
\State {\quad AxBxP-config = $\Omega_{c+p}$.pop()}
\State {\quad Convert-and-insert-AxBxP-Tensors ($DNN_{AxBxP}$,$W_{AxBxP}$,$A_{AxBxP}$, AxBxP-config)}
\State {\quad $Acc_{AxBxP}$ = evaluate($DNN_{AxBxP}$)}
\State {return AxBxP-config}
\end{algorithmic}
}
\end{algorithm}  

\begin{algorithm}
\caption{Convert-To-AxBxP}
\label{algo:axbxpRep}
\begin{algorithmic}[1]
\State \textbf{Input:}\{$X_{FxP}$: FxP tensor, ($K$, $N_{X}$, $\tilde{N}_X$): Ax-BxP configuration\}
\State \textbf{OUTPUT:} AxBxP tensor
\State { {\bf For each} scalar x in $X_{FxP}$}
\State { \quad $Block_{list}$ = Get-Significance-Sorted-Blocks (x, K, $N_{X}$) }
\State { \quad $\mathcal{I}_x$ = get-idx-first-Non-Zero-block ($Block_{list}$)}
\State {\quad $X_{AxFxP}$ = pick-insert-Blocks-in-Range ($\mathcal{I}_x$, $\mathcal{I}_x$-$\tilde{N}_X$)}
\State {return $X_{AxBxP}$}
\end{algorithmic}
\end{algorithm}  

\revise {We determine the best Ax-BxP configuration layer by layer for a given DNN as shown in Algorithm \ref{algo:Des_Ax-BxP_DNN} (lines 6-12). We start with a pre-trained $DNN_{FxP}$, where exact computations are performed in each layer. Once we find the best Ax-BxP configuration for a layer, we convert the operands of that layer to Ax-BxP format and proceed to the next layer. Algorithm \ref{algo:getBestConfig} describes the methodology to choose the best Ax-BxP configuration for a given DNN layer, block-size $K_{tgt}$ and target accuracy degradation $\gamma$. We set the operands to each of the Ax-BxP formats in $\Omega_{c+p}$ (lines 4-5), and evaluate the resulting $DNN_{AxBxP}$ (line 6). We perform the evaluation on a subset of the training dataset to speed-up the search. In our experiments with the ImageNet dataset, we perform re-training for 1 epoch on 120,000 images followed by testing on 5000 images to determine accuracy. Subsequently, we choose the first encountered Ax-BxP configuration that provides the desired accuracy (line 7).

We note that after the design space pruning described in section \ref{subsubsec:Prun}, the design space for finding the best approximation configuration is drastically reduced to $2,3$ and $5$ choices per layer for $K = 4, 3$ and $2$, respectively. Since we perform a greedy search on a layer-by-layer basis, the size of the pruned design space for a DNN with $N$ layers is at most $5N$. For $C$ choices, a DNN with $N$ layers and evaluation time of $T$ seconds per choice, the time taken to find the best Ax-BxP configuration is $C*N*T$.}

Algorithm \ref{algo:axbxpRep} outlines the pseudo code for converting FxP tensors to Ax-BxP tensors using the \emph{dynamic-idx} heuristic. It takes an FxP tensor ($X_{FxP}$) and Ax-BxP configuration ($K$, $N_{X}$, $\tilde{N}_X$) as inputs and produces an Ax-BxP tensor ($X_{AxBxP}$) as output. A key function of this algorithm is to determine indices ($I_{x}$) of the chosen blocks. To achieve this objective, we first convert fixed point scalars to blocked fixed-point scalars (line 2) and subsequently, pick $\tilde{N}_X$ contiguous blocks starting from the first non-zero block. Next, the chosen blocks and indices are inserted into the Ax-BxP tensor (line 4). After all scalars have been converted the $X_{AxBxP}$ tensor is returned (line 5).

\subsection{Ax-BxP DNN Accelerator}
\label{subsec:accdes}

\begin{wrapfigure}{r}{0.7\columnwidth}
    \vspace{-0.5em}
    \includegraphics[width=0.7\columnwidth]{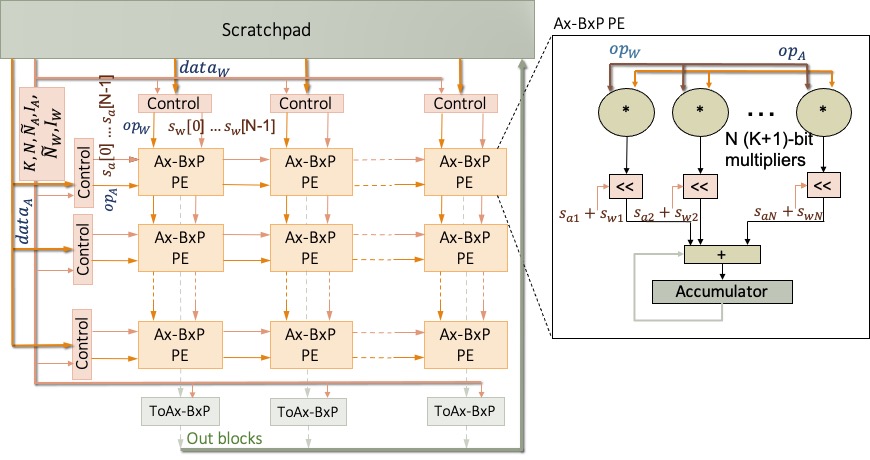}
    \caption{Ax-BxP accelerator architecture}
    \label{fig:Systolic_Array}
\end{wrapfigure}

Figure \ref{fig:Systolic_Array} shows the proposed Ax-BxP DNN accelerator, which is a conventional two-dimensional systolic-array based DNN accelerator with enhancements such as Ax-BxP PEs, control logic and peripheral units (ToAx-BxP) that support Ax-BxP computation. We design the Ax-BxP DNN accelerator for a fixed $K$ (although $K$ is a parameter in the RTL, it is specified at synthesis time). While the proposed approach can be applied with any DNN dataflow, for illustration we focus on the output stationary dataflow. The control logic partitions the Ax-BxP operands into blocks and determines the corresponding shift-amounts. These operand blocks and shift amounts are used to perform the Ax-BxP MAC operations in the Ax-BxP PEs. Finally, the output activations computed by the Ax-BxP PEs are converted into the Ax-BxP format by the ToAx-BxP logic, using the methodology described in Algorithm \ref{algo:axbxpRep} (see Section \ref{subsubsec:algorithms}). We now discuss the design of the \emph{Ax-BxP PE} and the \emph{Control} logic in detail. 

\subsubsection{Ax-BxP Processing Element (Ax-BxP PE)}
\label{subsec:pedes}

The Ax-BxP PEs contain $N = \ceil{8/K}$ signed multipliers of bitwidth $K+1$ and $N$ shifters to support the Ax-BxP computations. The partial products generated by the multipliers are shifted by the shift amounts determined by the control logic, and are accumulated at high-precision in the \finalrevise{32-bit} accumulator to generate the output activations. It is worth noting that as $L$ decreases, the throughput achieved by Ax-BxP PEs increases since a fixed number ($N$) of multiplications are performed in each cycle.

For a given $K$ and a given mode of Ax-BxP, Ax-BxP PEs support all the Ax-BxP configurations in $\Omega_{c+p}$ with block size $K$ by allowing different shift amounts. The different shift amounts are realized in the shifters using multiplexers. The number of multiplexers increases with the number of unique shift amounts to be supported, resulting in increases in energy and area overheads. It is straightforward to show that larger number of unique shift amounts must be supported by the Ax-BxP PE during dynamic Ax-BxP compared to static Ax-BxP. Therefore, the energy and area of Ax-BxP PEs are comparatively larger in the case of dynamic Ax-BxP.

\subsubsection{Control}
\label{subsec:controldes}

\newrevise{The control logic in the Ax-BxP DNN accelerator partitions each of the Ax-BxP operands $A_{AxBxP}$ and $W_{AxBxP}$ into signed blocks of $K+1$ bits, and determines the shift amounts corresponding to each of these blocks. The shift amounts are determined based on the indices of the operand blocks, where the shift amount corresponding to the $i^{th}$ block is computed as $i*K$. The indices of the operand blocks of $A_{AxBxP}$ ($W_{AxBxP}$) are determined from the parameters $\tilde{N}_A$ and $\mathcal{I}_A$ ($\tilde{N}_W$ and $\mathcal{I}_W$). Note that these parameters are obtained from the AxBxP format of the operands.

For a given layer of the DNN, in the static mode, the parameters $\tilde{N}_A$ and $\mathcal{I}_A$ ($\tilde{N}_W$ and $\mathcal{I}_W$) are fixed for all the scalar elements of $A_{AxBxP}$ ($W_{AxBxP}$) and therefore, are broadcast to the control blocks at the start of a layer's computations. As discussed in section \ref{subsubsec:designheuristics}, $\mathcal{I}_A$ ($\mathcal{I}_W$) is the index of the most significant block of $A_{AxBxP}$ ($W_{AxBxP}$), from which the indices of $\tilde{N}_A$ ($\tilde{N}_W$) consecutive blocks are derived. In the dynamic mode, the parameter $\tilde{N}_A$ ( $\tilde{N}_W$) is fixed while $\mathcal{I}_A$ ($\mathcal{I}_W$) varies within the operand tensor $A_{AxBxP}$ ($W_{AxBxP}$). Therefore, the parameter $\tilde{N}_A$ ( $\tilde{N}_W$) is broadcast to the control blocks at the start of a layer's computations. The parameters $\mathcal{I}_A$ and $\mathcal{I}_W$ are obtained from each scalar element of $A_{AxBxP}$ and $W_{AxBxP}$, respectively, since the indices of the chosen operand blocks are stored individually for every scalar element.}

It is worth noting that the design complexity of both the Ax-BxP PEs and the control logic depends on the number of unique shift amounts to be supported which in-turn depends on $L$. Therefore, constraining $L$ to be $\leq N$ minimizes the design complexity while still achieving good classification accuracy (as shown in section \ref{sec:Results}).

%% file: sections/exptsetup.tex
In this section, we present the experimental methodology used to evaluate Ax-BxP.

{\bf\noindent Accuracy Evaluation:} We evaluate Ax-BxP using three state-of-the-art image recognition DNNs for the ImageNet dataset, {\em viz.} ResNet50, MobileNetV2 and AlexNet. \revise{The ImageNet dataset has 1.28M training images and 50,000 test images. We use the entire test dataset for the accuracy evaluation. We perform upto 5 epochs of re-training for all the Ax-BxP configurations considered. We use an Intel Core i7 system with NVIDIA GeForce 2080 (Turing) graphics card for the simulations.}

{\bf\noindent System Energy and Performance Evaluation:} We design the Ax-BxP DNN accelerator by expanding the conventional systolic array accelerator modelled in ScaleSim \cite{samajdar2018scale} to include enhancements such as Control logic, ToAxBxP logic and the AxBxP PEs, all synthesized to the 15nm technology node using Synopsys Design Compiler. We consider an output-stationary systolic array of size 32x32 and \revise{on-chip scratch-pad memory of 2MB, operating at 1GHz}. The on-chip memory is modelled using CACTI \cite{muralimanohar2009cacti}. We design our baseline FxP8 accelerator, also synthesized to the 15nm technology node using Synopsys Design Compiler, as a conventional systolic array with FxP8 PEs that can implement 8-bit MAC operations. The system-level energy and performance benefits of the Ax-BxP accelerator are evaluated against the FxP8 baseline. \revise{To evaluate the benefits over a mixed-precision baseline, we adopt the HAQ \cite{wang2019haq,HAQ_impl} precision configuration and design a power-gated FxP8 baseline with power-gated FxP8 PEs, i.e, we design the PEs for a maximum precision of 8 bits and power-gate the unused portions during lower-precision computations. Additionally, we evaluate the benefits of the proposed Ax-BxP in the Bit-Fusion accelerator \cite{sharma2018bit}.} 


%% file: sections/results.tex
\noindent In this section, we demonstrate the energy and performance benefits of Ax-BxP at the processing element (PE) level and system level using the proposed Ax-BxP DNN accelerator. We also demonstrate the benefits of Ax-BxP in the Bit-Fusion \cite{sharma2018bit} accelerator. 

\subsection{PE-level energy and area benefits}
\label{subsec:PE_Ben}

\begin{figure}[htb]
  \centering
  \includegraphics[width=0.7\columnwidth]{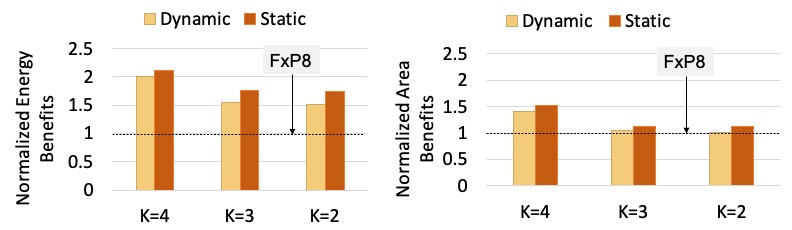}
  \vspace*{-6pt}
  \caption{Computation energy and area benefits over an FxP8 baseline}
  \label{fig:PEEnergyArea}
\end{figure}

Figure \ref{fig:PEEnergyArea} shows the energy and area benefits of the Ax-BxP PE for $K=2,3,4$ when compared to an 8-bit fixed-point (FxP8) PE. On average, the energy and area benefits of the Ax-BxP PE in dynamic mode are \revise{1.69x and 1.12x}, respectively. In static mode, the average energy and area benefits are \revise{1.87x and 1.25x}, respectively.

\begin{figure}[htb]
  \centering
  \includegraphics[width=\columnwidth]{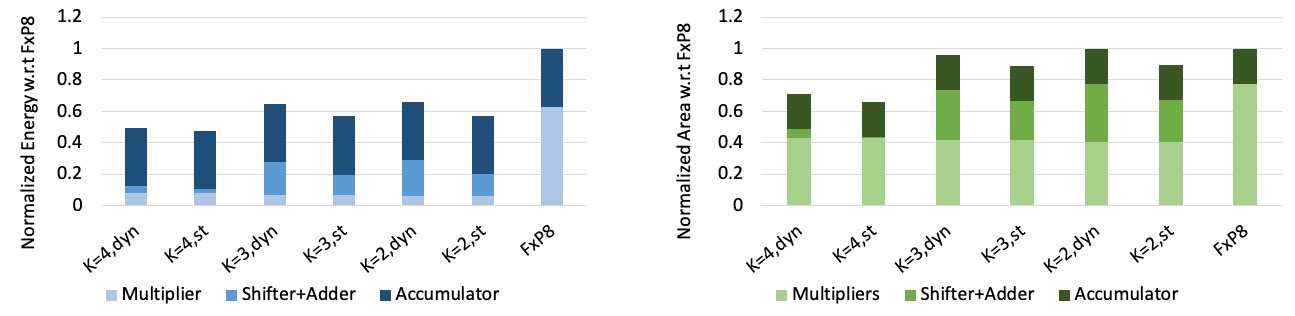}
  \vspace*{-4pt}
  \caption{Computation energy breakdown (left) and area breakdown (right) for Ax-BxP and the FxP8 baseline}
  \label{fig:PEBreakdown}
\end{figure}

Recall from section \ref{subsec:accdes} that for a given $K$, the Ax-BxP PEs can support any Ax-BxP configuration by allowing shifts by different amounts. The energy and area overheads of Ax-BxP PEs increase with an increase in the number of unique shift amounts to be supported. The number of unique shift amounts to be supported is proportional to $N=\ceil{8/K}$. Therefore the energy benefits decrease as $K$ decreases for both static and dynamic modes of operation. Furthermore, since greater numbers of shift amounts need to be supported for the dynamic mode, the energy benefits during dynamic Ax-BxP are lower than static Ax-BxP for all $K$. Figure \ref{fig:PEBreakdown} (left) shows the energy breakdown of multipliers, shifters and adders, and the accumulator in Ax-BxP PEs for all $K$ in both static and dynamic modes of Ax-BxP. For a given value of $K$, we observe that the overhead due to re-configurability (i.e shifters and adders) is greater in dynamic mode vs. static mode. \revise{Figure \ref{fig:PEBreakdown} (right) shows the area breakdown of the Ax-BxP PEs and the FxP8 PE. Similar to energy, we find that for a given $K$, the re-configurability overhead is greater in dynamic vs static mode. The energy and area overheads of re-configurability are larger for smaller $K$ because of the greater number of shift amounts to be supported.}

\subsection{System Benefits and  Accuracy of dynamic Ax-BxP with network-level precision re-configuration}
\label{subsec:DynBenefitsAcrossNetworks}

\begin{figure}[htb]
  \vspace*{-8pt}
  \centering
  \includegraphics[width=\columnwidth]{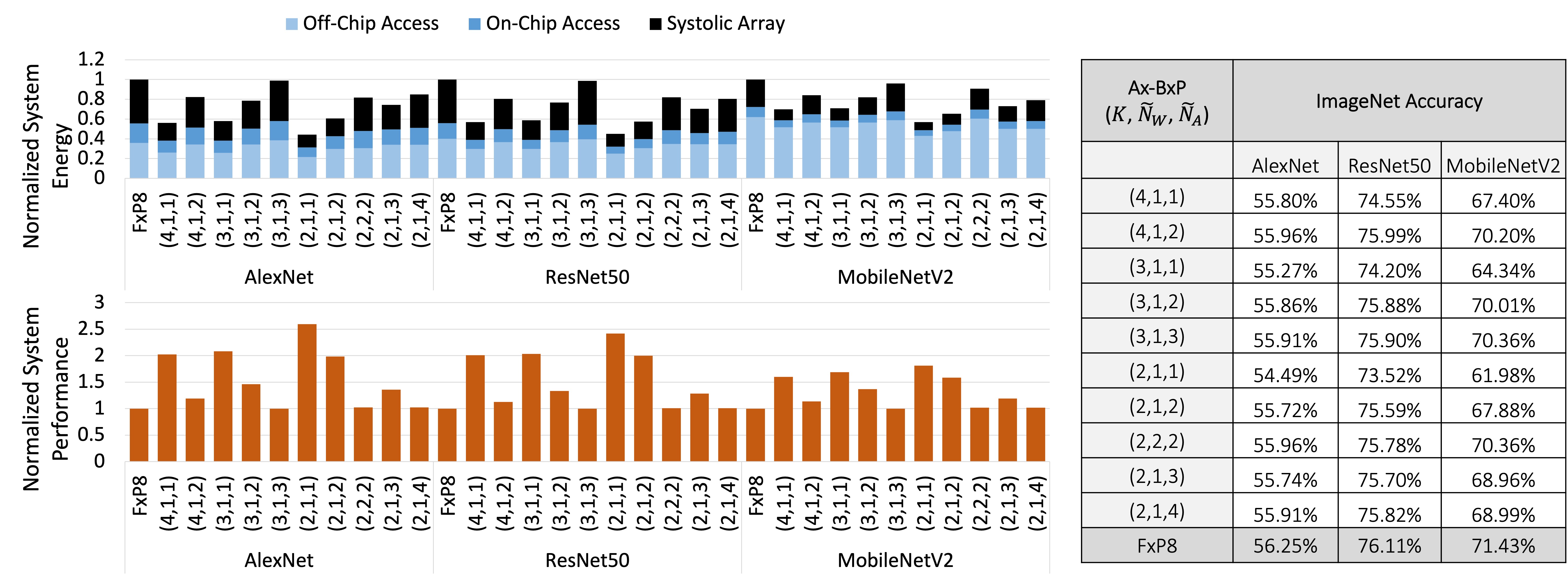}
  \vspace*{-6pt}
  \caption{System-level Benefits with dynamic Ax-BxP}
  \label{fig:SysDyn}
  \vspace*{-8pt}
\end{figure}

Figure \ref{fig:SysDyn} shows the system-level benefits of dynamic Ax-BxP inference compared to the FxP8 baseline at iso-area. In this experiment, for a given network, we maintain the Ax-BxP configurations uniform across its layers. We find that the best Ax-BxP configuration in $\Omega_{c+p}$ for a given $K$, \emph{i.e} the configuration that provides maximum energy benefits with small accuracy loss ($\sim 1\%$), varies across the networks considered. 

\begin{table}[htb]
    \centering
    \caption{Best Dynamic Ax-BxP configurations}
    \vspace{-10pt}
    \begin{tabular}{|c|c|c|c|}
    \hline
         Block-size & AlexNet & ResNet50 & MobileNetV2 \\
         \hline
         $K=4$ & (4,1,1) & (4,1,2) & (4,1,2) \\ 
         $K=3$ & (3,1,1) & (3,1,2) & (3,1,2) \\
         $K=2$ & (2,1,2) & (2,1,2) & (2,2,2) \\
         \hline
    \end{tabular}
    \label{tab:BestConfig}
\end{table}

Table \ref{tab:BestConfig} shows the best Ax-BxP configuration for a each $K$ and each network considered. The proposed Ax-BxP DNN accelerator efficiently supports varying  configurations across networks and achieves system-level energy \revise{reduction of 1.19x-1.78x, 1.01x-1.72x and 1.1x-2.26x for $K=$ 4, 3 and 2, respectively compared to the FxP8 baseline at iso-area. Figure \ref{fig:SysDyn} shows the breakdown of the normalized system-energy into Off-Chip memory access energy, On-Chip Buffer access energy and the Systolic-Array energy. We achieve a significant reduction in each of these components in our proposed Ax-BxP accelerator across DNNs and across the approximation configurations considered.}

\revise{The 1.04x-3.41x reduction in the systolic-array energy stems primarily from the Ax-BxP PE benefits discussed in Section \ref{subsec:PE_Ben}. Additionally, for a given $K$, when $L$ decreases, the throughput of the Ax-BxP DNN accelerator increases as discussed in Section \ref{subsec:accdes}. Therefore, the overall inference cycles reduces, resulting in further reduction in the systolic array energy.  However, the Imagenet classification accuracy decreases with smaller $L$. Furthermore, by selecting to store a subset of the operand blocks, we achieve 1.01x-1.6x and 1.03x-2.17x reduction in Off-chip and On-chip memory access energy, respectively. For a given $K$, the memory-access energy decreases as $L$ decreases. This is because of the reduction in memory footprint and the number of accesses despite the overhead of storing the $\ceil{log2((N-\tilde{N}) + 1)}$ bits of operand block indices during dynamic Ax-BxP.}  



The performance benefits during dynamic Ax-BxP compared to FxP8 at iso-area are also shown in Figure \ref{fig:SysDyn}. We obtain 1.13x-2.02x, 1.01x-2.08x and 1.02x-2.59x performance benefits for $K=$ 4,3 and 2, respectively for the configurations listed in Table \ref{tab:BestConfig}. For a given $K$, smaller $L$ results in increased throughput, which in turn results in increased performance.


\subsection{System-level energy benefits and accuracy with intra-network precision reconfiguration}
\label{subsec:variablesysEnergy}

\begin{figure}[htb]
  \vspace*{-0pt}
  \centering
  \includegraphics[width=\columnwidth]{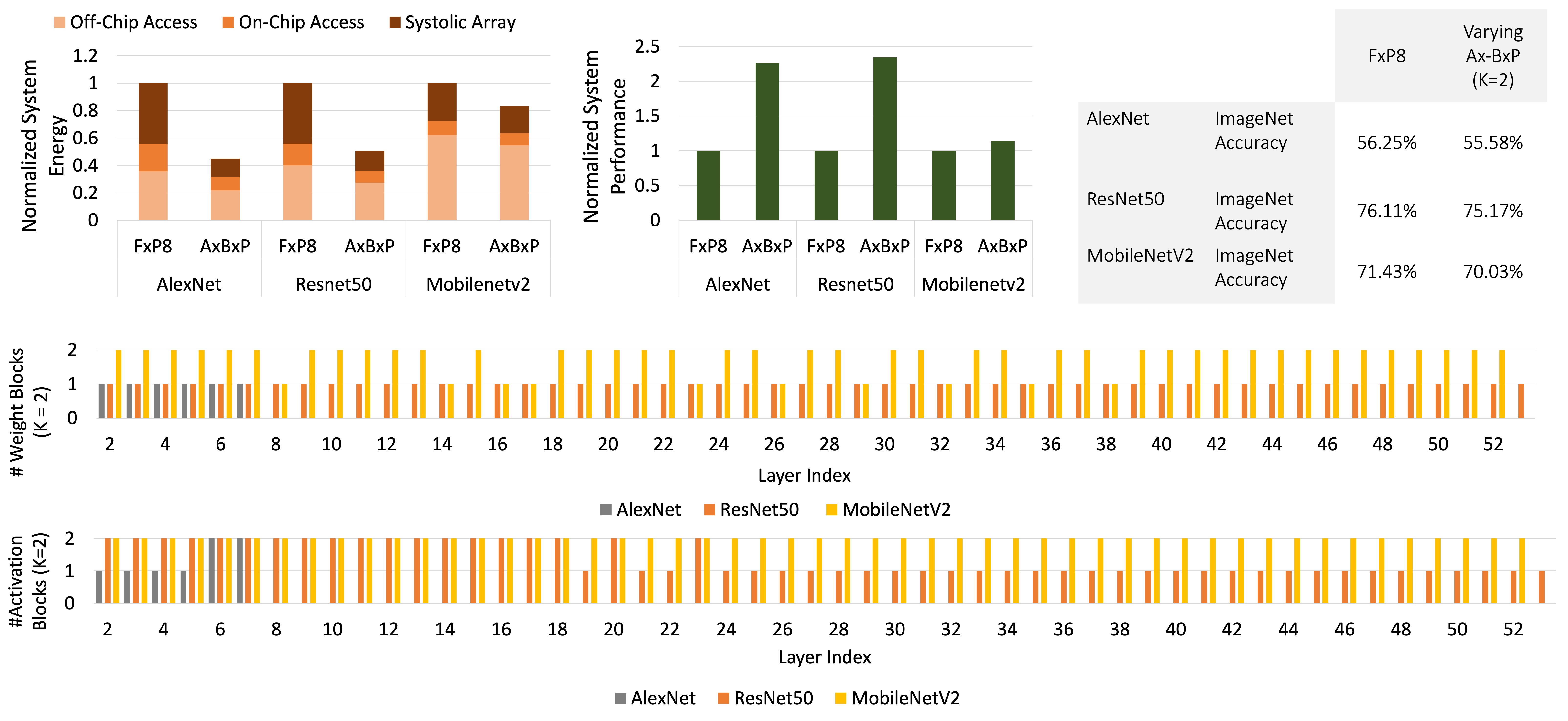}
  \vspace*{-15pt}
  \caption{System-level benefits and ImageNet accuracy for mixed-precision networks}
  \label{fig:VarResult}
  \vspace*{-0pt}
\end{figure}

The benefits of Ax-BxP can be further highlighted in the context of variable-precision DNNs that have different layer-wise precision requirements. The uniform Ax-BxP configuration of $(2,1,1)$ provides the maximum system-level energy benefits. However, it suffers from significant accuracy degradation. To improve the classification accuracy and to maximize the system-benefits, we vary the precision in a coarse-grained manner across the layers of these DNNs with $K=2$. We use the Ax-BxP configurations $(2,1,2)$ and $(2,1,1)$ for ResNet50 and AlexNet, and $(2,2,2)$ and $(2,1,2)$ for MobileNetV2. The layer-wise precision configurations considered are shown in Figure \ref{fig:VarResult}. \finalrevise{Following previous studies \cite{hubara2017quantized,zhou2016dorefanet,choi2018pact}, we do not approximate the first and last layer computations as they have been shown to impact the classification accuracy significantly}. We achieve \revise{1.12x-2.23x} reduction in system energy and \revise{1.13x-2.34x} improvement in system performance compared to the FxP8 baseline with small loss in classification accuracy. 

\begin{figure}[htb]
  \vspace*{-0pt}
  \centering
  \includegraphics[width=
  \columnwidth]{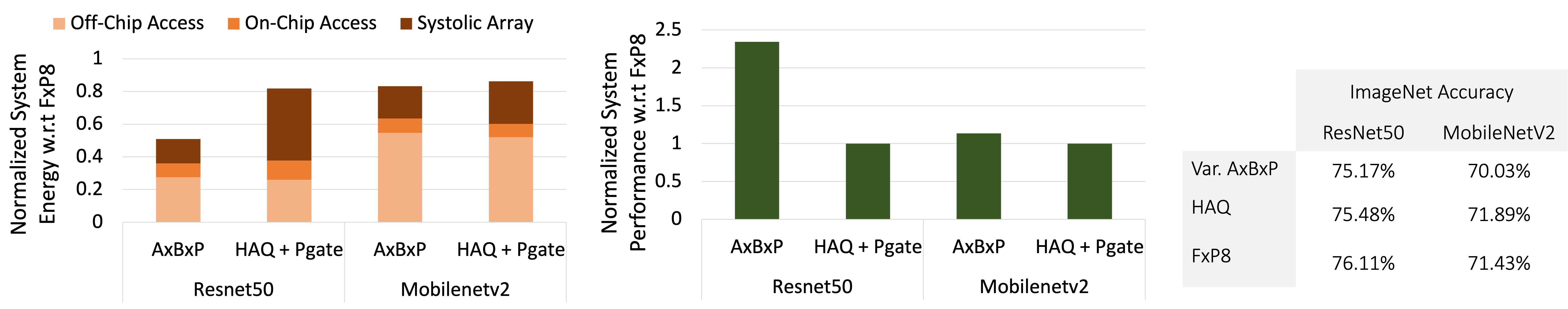}
  \vspace*{-10pt}
  \caption{Comparison to HAQ precision configuration implemented in a power-gated FxP8 systolic array}
  \label{fig:HAQ_Comp}
  \vspace*{-0pt}
\end{figure}

\revise{Next, we compare Ax-BxP against the HAQ \cite{wang2019haq,HAQ_impl} mixed-precision configuration implemented on a conventional systolic array accelerator with power-gated PEs. We design the PEs to support the worst-case precision of 8-bits and power-gate the unused portions during sub-8-bit computations. Figure \ref{fig:HAQ_Comp} shows the normalized system-energy breakdown, performance, and ImageNet accuracy of Ax-BxP and HAQ implementations. We observe that Ax-BxP achieves 1.04x-1.6x system-energy reduction and 1.1x-2.34x performance improvement compared to HAQ. The memory access energy (off-chip + on-chip) of AxBxP is 0.95x and 1.06x of the HAQ memory-access energy for ResNet50 and MobileNetV2, respectively. The small overhead in case of MobileNetV2 is caused by storing the Ax-BxP operand block indices. Despite this overhead, Ax-BxP achieves superior system benefits compared to HAQ by substantially reducing the systolic-array energy by 1.32x-2.95x. This is the result of lower overall inference cycles for Ax-BxP, achieved through superior systolic-array utilization compared to the power-gated FxP8 implementation of HAQ. As shown in Figure \ref{fig:HAQ_Comp}, HAQ does not provide any performance improvement compared to FxP8 because the power-gated PEs cannot increase the throughput of the systolic-array.}


\subsection{Benefits of Ax-BxP in the Bit-Fusion accelerator}
\label{subsec:bitfusionBenefits}
\begin{figure}[htb]
  \centering
  \includegraphics[width=0.7\columnwidth]{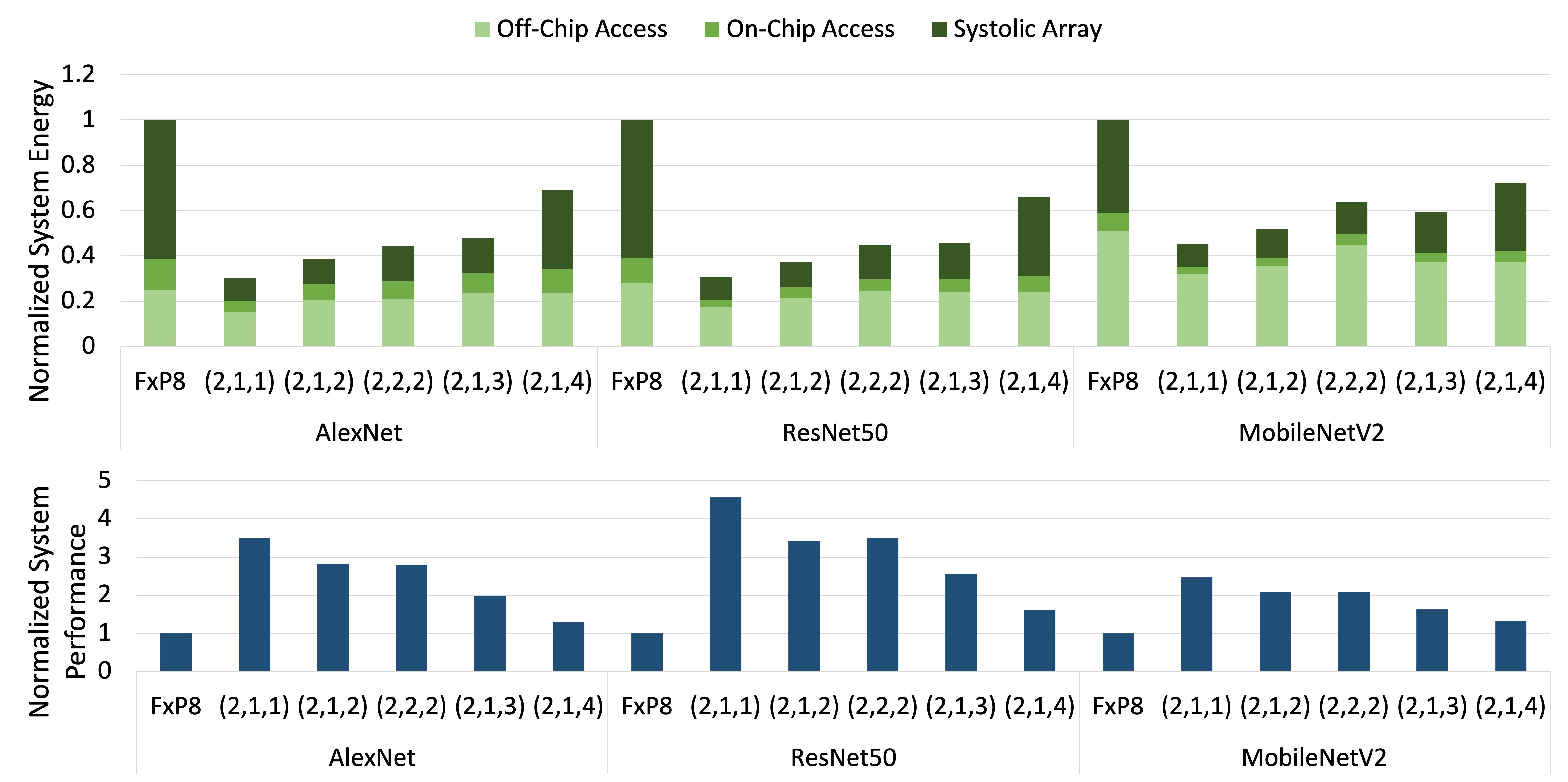}
  \caption{Ax-BxP Benefits in BitFusion}
  \label{fig:BitFusionBenefits}
\end{figure}

To demonstrate the broad applicability of Ax-BxP, we implement it on top of the
Bit-Fusion accelerator architecture~\cite{sharma2018bit}. The energy and performance benefits of dynamic Ax-BxP in the Bit-Fusion accelerator compared to exact FxP8 computations are shown in Figure \ref{fig:BitFusionBenefits}. We have considered the Ax-BxP configurations with $K=2$, since the bit-bricks in Bit-Fusion PE are designed for a block-size of $2$. By performing approximations using Ax-BxP, we could achieve energy benefits of upto \revise{3.3x} and performance benefits of upto \revise{4.6x} in the Bit-Fusion accelerator. The Bit-Fusion PEs achieve a comparatively higher throughput for a given $L$. The maximum increase in throughput is 16x when $L=1$, resulting in the 4.6x benefits for the configuration $(2,1,1)$. 

\begin{figure}[htb]
  \vspace*{-0pt}
  \centering
  \includegraphics[width=0.7\columnwidth]{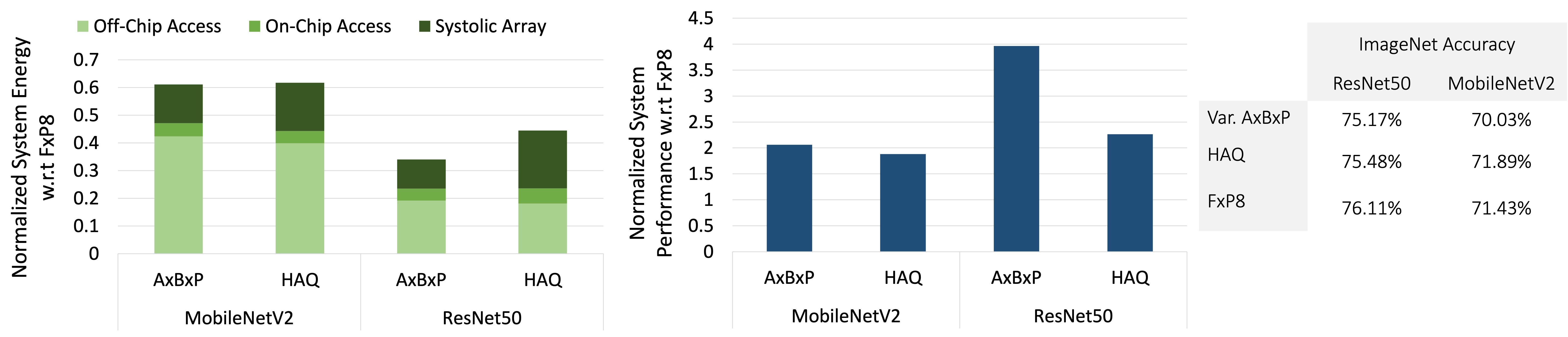}
  \vspace*{-10pt}
  \caption{\newrevise{Comparison of AxBxP and HAQ when implemented in the BitFusion Accelerator}}
  \label{fig:HAQ_Bf}
  \vspace*{-0pt}
\end{figure}

\newrevise{We further evaluate the benefits of Ax-BxP over HAQ when both are implemented on the Bitfusion accelerator. As shown in Figure \ref{fig:HAQ_Bf}, Ax-BxP achieves 1.01x-1.31x reduction in system energy and  1.09x-1.75x improvement in system performance compared to HAQ. Similar to the discussion in section \ref{subsec:variablesysEnergy}, the small overhead in the memory access energy (on-chip + off-chip) of Ax-BxP compared to HAQ in MobileNetV2 (1.06x) can be attributed to the overhead in storing the Ax-BxP operand block-indices. However, Ax-BxP achieves 1.25x to 1.98x reduction in systolic-array energy, which results in an overall reduction in system energy compared to HAQ. Further, Ax-BxP  achieves higher throughput than HAQ as a result of superior utilization of the on-chip bit-bricks (2-bit signed multipliers) of the Bitfusion accelerator. For instance, in layers 42 to 52 of MobileNetV2, HAQ \cite{wang2019haq} multiplies 6-bit weigths with 6-bit activations, requiring 9 bit-bricks whereas Ax-BxP (Figure \ref{fig:VarResult}) performs these computations using 2 blocks of weights and 2 blocks of activations, requiring only 4 bit-bricks. By dynamically choosing the weight and activation blocks, Ax-BxP minimizes the accuracy degradation (shown in Figure \ref{fig:VarResult}).}


\subsection{System benefits and  accuracy of static Ax-BxP}
\label{subsec:StBenefitsAcrossNetworks}

\begin{figure}[htb]
  \vspace*{-4pt}
  \centering
  \includegraphics[width=\columnwidth]{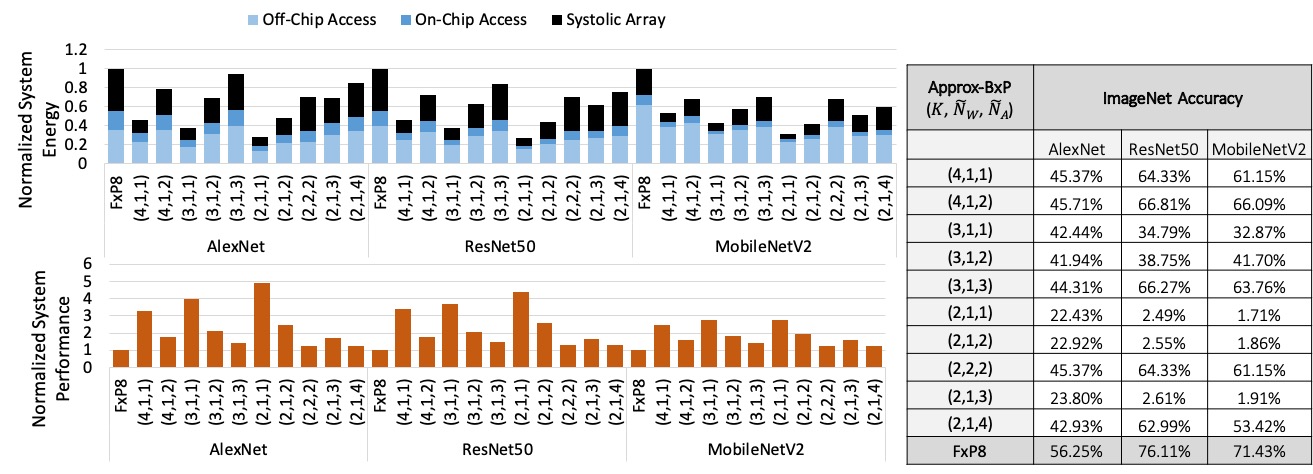}
  \vspace*{-15pt}
  \caption{System-level Benefits with static Ax-BxP}
  \label{fig:SysSt}
  \vspace*{-4pt}
\end{figure}

Figure \ref{fig:SysSt} shows the system benefits of static Ax-BxP compared to the FxP8 baseline at iso-area. The energy benefits during static Ax-BxP are greater than the dynamic Ax-BxP across networks and across configurations. This is because the Ax-BxP PEs are simpler in terms of the number of shift amounts to be supported during the static mode, compared to the dynamic mode. Furthermore, the memory footprint of the operands are lower during the static mode compared to the dynamic mode, since the cost of storing and fetching $\mathcal{I}_W$ and $\mathcal{I}_A$ are amortized across the tensors $W$ and $A$, respectively. As a result the memory access energy is lower and the compute energy is lower during static Ax-BxP. However, for equal re-training effort (5 epochs), the ImageNet accuracy degradation with static Ax-BxP is significantly higher than dynamic Ax-BxP across configurations and networks. The performance benefits with static Ax-BxP are also shown in Figure \ref{fig:SysSt}. The performance benefits during static Ax-BxP are greater than dynamic Ax-BxP across networks and Ax-BxP configurations. This is because in static mode, the Ax-BxP PEs are smaller, which is exploited in the iso-area design to achieve higher throughput. 

%% file: sections/relatedWork.tex
The high computation and storage demands posed by DNNs have motivated several efforts to focus on precision scaling. Many of the early efforts  \cite{zhou2016dorefanet, li2016ternary, hubara2017quantized, courbariaux1602binarynet, rastegari2016xnor} on precision scaling are effective in small networks, but they suffer significant accuracy degradation in large networks. 

More recent efforts \cite{choi2018pact,zhou2017bq,mishra2017wrpn,han2015deep,jain2019biscaled,jain2018compensated} have developed advanced quantization techniques and training methodologies that work well for a wide range of networks and effectively reduce the bit-widths of data-structures to below 8 bits. Notably, PACT~\cite{choi2018pact} has demonstrated successful inference using only 2-bit precision for weights and activations, except in the first and last layers, which are evaluated at 8-bit precision. Other efforts such as BQ \cite{zhou2017bq} and WRPN \cite{mishra2017wrpn} also achieve good inference accuracy using 2-bit weights and activations, by using techniques such as Balanced Quantization and Model scaling, respectively. Deep Compression \cite{han2015deep} employs a combination of pruning, quantization and Huffman coding to reduce the model size. Bi-Scaled DNN \cite{jain2019biscaled} and Compensated DNN \cite{jain2018compensated} leverage the value statistics in DNNs and design number-formats and error compensation schemes that effectively reduce the overall bit-width. Although these efforts enable DNN inference with ultra-low (below 8-bit) precision, a common precision is not optimal across networks or even across layers within a network. For example, it is a common practice to retain the first and last layers at higher precision while quantizing intermediate layers to very low precision in order to preserve accuracy. This trend is carried further by works like HAQ \cite{wang2019haq}, which demonstrate that the minimum precision requirement varies within a network, across different layers.  

Since varying precision requirements are inherent across DNNs, recent efforts~\cite{umuroglu2018bismo, judd2016stripes, sharma2018bit} have focused on the design of precision re-configurable hardware. BISMO \cite{umuroglu2018bismo} proposes a parallelized bit-serial architecture that offers maximum flexibility in terms of the bit-widths it can support. Stripes \cite{judd2016stripes} is a similar work that uses bit-serial hardware design to support variable precision computation. Albeit offering maximum flexibility, the performance of bit-serial hardware is limited by high latency and energy caused by its serial nature. On the other hand, fixed-precision hardware needs to be designed to support maximum precision and hence is over-designed for applications with low precision requirement. Instead of performing computations serially at the granularity of bits, BitFusion \cite{sharma2018bit} explores serial computation at the granularity of a group of bits and demonstrates superior energy benefits compared to state-of-the-art bit-serial and fixed-precision accelerators. However, \emph{none of these efforts explore the use of approximations to improve the efficiency of variable-precision hardware}. Exploiting the resilience of DNNs to approximations, we propose approximate blocked computation as a next step that is complementary to previous efforts. Computations are performed block-wise, where blocks are a group of bits of fixed length. Our approximation methodology reduces the number of block-wise computations that need to be performed, while maintaining the inference accuracy.


The design of approximate multipliers has been extensively explored in the literature. These efforts can be broadly classified into three categories -- efforts that focus on the design of general-purpose approximate circuits, efforts that approximate partial-product accumulation, and efforts that approximate partial product generation. Efforts such as \cite{VOS_DSP, Salsa, EvolApproach, MACACO} focus on general-purpose approximate circuit design using voltage over-scaling and logic simplification techniques. The logic simplification approach used in these methods eliminates transistors or gates to trade off accuracy for efficiency. A more structured way to approximate multiplier circuits is to systematically minimize the computation and accumulation of partial products.


Energy reduction during partial product accumulation can be achieved by approximate adders \cite{SpecAdd, Truncation-ErrTolerant, IMPACT} or by approximate accumulation techniques \cite{LogicCompr}. While these efforts focus on minimizing the energy consumption of accumulation, we note that the multipliers are the primary sources of energy consumption during MAC operations. Several previous works \cite{Kulkarni,InAccCounter_Lin,LogApprox,AlphabetSet,AlphSet_DSP,Yeo} have explored the approximation of partial product generation.


In \cite{Kulkarni}, the authors propose a 2x2 under-designed multiplier block and build arbitrarily large power efficient in-accurate multipliers. The inaccurate 4:2 counter in \cite{InAccCounter_Lin} can effectively reduce the partial product stages of the Wallace multiplier. In \cite{LogApprox}, the authors substitute multiplication with additions and shift operations by representing the integer operands as logarithms with an error correction factor. The computation sharing multiplier proposed in \cite{AlphabetSet,AlphSet_DSP} specifically targets computation re-use in vector-scalar products. Reference \cite{Yeo} proposes an approximate multiplier that performs approximation on only the multiplication of lower-order bits. These efforts achieve only computational energy benefits. In contrast, since our proposed Ax-BxP method minimizes the number of operand blocks used in computation, we achieve savings in terms of memory footprint and memory traffic in addition to computational energy savings. Other efforts such as \cite{TrucSwartzlander,AreaEffGuibaly,FixWidthChen,Narayanmoorthy,DRUM,PP_Perf} have taken a similar approach.

Operand bit-width truncation to minimize partial product generation is explored by efforts such as \cite{TrucSwartzlander}, \cite{AreaEffGuibaly} and \cite{FixWidthChen}. However, these efforts exhibit poor performance during small bit-width computations. In \cite{Narayanmoorthy}, the authors extract an m-bit segment from an n-bit operand and perform an m-bit (m<n) bit multiplication, achieving significant energy benefits. However the segments must by atleast n/2 bits long, thus limiting the energy savings. An improvement over \cite{Narayanmoorthy} is proposed by \cite{DRUM} that reduces the segment size beyond n/2 while minimizing error, enabling dynamic range multiplication. However, this approach involves complex-circuitry such as leading one-bit detectors, barrel shifters etc., which introduces considerable delay, area and energy overheads that decrease the approximation benefits. The partial product perforation method proposed by \cite{PP_Perf} aims at generating fewer partial products by dropping a few bits of one operand during multiplication. Since this approach reduces the bit-width (precision) of just one operand, it does not fully utilize the benefits of precision scaling. Moreover, it requires complex error correction methods that further limit the benefits of approximation method. 

Additionally, none of the approximation efforts discussed thus far have focused on variable precision computations which we have explored in our work. Reference \cite{RedPrec_FP} proposes dynamic range floating-point (FP) format to support varying precision. However, the area and power cost of supporting FP computations is much higher than fixed-point (FxP) computations. In \cite{jain2018compensated}, the authors propose error-compensation techniques for reduced-precision FxP multiplication. A novel number format to represent dual-precision FxP numbers is proposed in \cite{jain2019biscaled}. Our proposed Ax-BxP format supports a wide range of precision requirements. It enables efficient re-configurability at the block granularity while minimizing the approximation errors.

%% file: sections/conclusion.tex
{\noindent}
Efforts to design ultra-low precision DNNs suggest that the minimum bit-width requirement varies across and within DNNs. Optimally supporting such varying precision configurations in DNN accelerators is therefore a key challenge.
We address this challenge algorithmically and in hardware using our proposed Approximate Blocked Computation method, Ax-BxP. We demonstrate the effectiveness of Ax-BxP in achieving good classification accuracy on the ImageNet dataset with state-of-the art DNNs such as AlexNet, ResNet50 and MobileNetV2. Ax-BxP provides upto \revise{1.74x and 2x} benefits in system energy and performance respectively, while varying configurations across networks. Further, with varying configurations at the layer level, Ax-BxP  achieves upto \revise{2.23x and 2.34x} improvements in system energy and performance, respectively.

%% file: paper.bbl
\begin{thebibliography}{10}

\bibitem{AlphaGO}
J.~X. {Chen}.
\newblock The evolution of computing: Alphago.
\newblock {\em Computing in Science Engineering}, 18(4):4--7, 2016.

\bibitem{GPT3}
Dario Amodei et~al. T.~Brown.
\newblock Language models are few-shot learners.
\newblock 2020.

\bibitem{Eff-Net}
Mingxing Tan and Quoc Le.
\newblock {E}fficient{N}et: Rethinking model scaling for convolutional neural
  networks.
\newblock In {\em Proceedings of the 36th International Conference on Machine
  Learning}, pages 6105--6114, 2019.

\bibitem{venkataramani2016}
S.~{Venkataramani}, K.~{Roy}, and A.~{Raghunathan}.
\newblock Efficient embedded learning for iot devices.
\newblock In {\em Asia and South Pacific Design Automation Conference
  (ASP-DAC)}, pages 308--311, Jan 2016.

\bibitem{sze2017}
V.~{Sze}, Y.~{Chen}, T.~{Yang}, and J.~S. {Emer}.
\newblock Efficient processing of deep neural networks: A tutorial and survey.
\newblock {\em Proceedings of the IEEE}, 105(12):2295--2329, Dec 2017.

\bibitem{TPU}
Doe Hyun et~al. N.~P.~Jouppi.
\newblock In-datacenter performance analysis of a tensor processing unit.
\newblock {\em SIGARCH Comput. Archit. News}, 45(2), 2017.

\bibitem{wang2019haq}
Kuan Wang, Zhijian Liu, Yujun Lin, Ji~Lin, and Song Han.
\newblock Haq: Hardware-aware automated quantization with mixed precision.
\newblock In {\em Proc. CVPR}, 2019.

\bibitem{mishra2017wrpn}
Asit Mishra, Eriko Nurvitadhi, Jeffrey~J Cook, and Debbie Marr.
\newblock Wrpn: wide reduced-precision networks.
\newblock 2017.

\bibitem{zhou2017bq}
Shuchang Zhou, Yuzhi Wang, He~Wen, Qinyao He, and Yuheng Zou.
\newblock Balanced quantization: An effective and efficient approach to
  quantized neural networks.
\newblock {\em JCST}, 32(4), 2017.

\bibitem{choi2018pact}
Jungwook Choi, Zhuo Wang, Swagath Venkataramani, Pierce I-Jen Chuang,
  Vijayalakshmi Srinivasan, and Kailash Gopalakrishnan.
\newblock Pact: Parameterized clipping activation for quantized neural
  networks.
\newblock 2018.

\bibitem{jain2019biscaled}
S.~{Jain}, S.~{Venkataramani}, V.~{Srinivasan}, J.~{Choi}, K.~{Gopalakrishnan},
  and L.~{Chang}.
\newblock Biscaled-dnn: Quantizing long-tailed datastructures with two scale
  factors for deep neural networks.
\newblock In {\em 56th ACM/IEEE Design Automation Conference (DAC)}, pages
  1--6, 2019.

\bibitem{Proteus}
Patrick Judd, Jorge Albericio, Tayler Hetherington, Tor~M. Aamodt,
  Natalie~Enright Jerger, and Andreas Moshovos.
\newblock Proteus: Exploiting numerical precision variability in deep neural
  networks.
\newblock In {\em In Proc. ICS}, 2016.

\bibitem{PrecisionQuant}
Soheil Hashemi, Nicholas Anthony, Hokchhay Tann, R.~Iris Bahar, and Sherief
  Reda.
\newblock Understanding the impact of precision quantization on the accuracy
  and energy of neural networks.
\newblock In {\em In Proc. DATE}, 2017.

\bibitem{umuroglu2018bismo}
Yaman Umuroglu, Lahiru Rasnayake, and Magnus Sjalander.
\newblock Bismo: A scalable bit-serial matrix multiplication overlay for
  reconfigurable computing.
\newblock In {\em ICFPL}. IEEE, 2018.

\bibitem{judd2016stripes}
P.~{Judd}, J.~{Albericio}, T.~{Hetherington}, T.~M. {Aamodt}, and
  A.~{Moshovos}.
\newblock Stripes: Bit-serial deep neural network computing.
\newblock In {\em MICRO}. IEEE, 2016.

\bibitem{loom}
Sayeh Sharify, Alberto~Delmas Lascorz, Kevin Siu, Patrick Judd, and Andreas
  Moshovos.
\newblock Loom: Exploiting weight and activation precisions to accelerate
  convolutional neural networks.
\newblock In {\em DAC}. ACM, 2018.

\bibitem{sharma2018bit}
Hardik Sharma, Jongse Park, Naveen Suda, Liangzhen Lai, Benson Chau, Joon~Kyung
  Kim, Vikas Chandra, and Hadi Esmaeilzadeh.
\newblock Bit fusion: Bit-level dynamically composable architecture for
  accelerating deep neural networks.
\newblock In {\em Proc. ISCA}, 2018.

\bibitem{Reconf_JETACS}
V.~{Camus}, L.~{Mei}, C.~{Enz}, and M.~{Verhelst}.
\newblock Review and benchmarking of precision-scalable multiply-accumulate
  unit architectures for embedded neural-network processing.
\newblock {\em IEEE Journal on Emerging and Selected Topics in Circuits and
  Systems}, 9(4):697--711, 2019.

\bibitem{axnn}
Swagath Venkataramani, Ashish Ranjan, Kaushik Roy, and Anand Raghunathan.
\newblock Axnn: Energy-efficient neuromorphic systems using approximate
  computing.
\newblock In {\em Proceedings of the 2014 International Symposium on Low Power
  Electronics and Design}, page 27–32, 2014.

\bibitem{chippa-asilomar13}
Vinay~K. Chippa, Swagath Venkataramani, Srimat~T. Chakradhar, Kaushik Roy, and
  Anand Raghunathan.
\newblock Approximate computing: An integrated hardware approach.
\newblock In {\em 2013 Asilomar Conference on Signals, Systems and Computers},
  pages 111--117, 2013.

\bibitem{DigSerial}
R.I. Hartley and K.K. Parhi.
\newblock {\em Digit-Serial Computation}.
\newblock 1995.

\bibitem{zhou2016dorefanet}
S.Zhou et~al.
\newblock Dorefa-net: Training low bitwidth convolutional neural networks with
  low bitwidth gradients.
\newblock {\em arXiv preprint arXiv:1606.06160}, 2016.

\bibitem{samajdar2018scale}
Ananda Samajdar, Yuhao Zhu, Paul Whatmough, Matthew Mattina, and Tushar
  Krishna.
\newblock Scale-sim: Systolic cnn accelerator.
\newblock 2019.

\bibitem{muralimanohar2009cacti}
Muralimanohar Naveen; Balasubramonian Rajeev; Jouppi~Norman P.
\newblock Cacti 6.0: A tool to model large caches.
\newblock Technical report, 2009.

\bibitem{HAQ_impl}
Kuan Wang, Zhijian Liu, Yujun Lin, Ji~Lin, and Song Han.
\newblock https://github.com/mit-han-lab/haq.

\bibitem{hubara2017quantized}
Itay Hubara, Matthieu Courbariaux, Daniel Soudry, Ran El-Yaniv, and Yoshua
  Bengio.
\newblock Quantized neural networks: Training neural networks with low
  precision weights and activations.
\newblock {\em JMLR}, 18(1), 2017.

\bibitem{li2016ternary}
Fengfu Li, Bo~Zhang, and Bin Liu.
\newblock Ternary weight networks.
\newblock 2016.

\bibitem{courbariaux1602binarynet}
Matthieu Courbariaux, Itay Hubara, Daniel Soudry, Ran El-Yaniv, and Yoshua
  Bengio.
\newblock Binarynet: Training deep neural networks with weights and activations
  constrained to+ 1 or- 1.
\newblock 2016.

\bibitem{rastegari2016xnor}
Mohammad Rastegari, Vicente Ordonez, Joseph Redmon, and Ali Farhadi.
\newblock Xnor-net: Imagenet classification using binary convolutional neural
  networks.
\newblock In {\em Computer Vision -- ECCV}. Springer, 2016.

\bibitem{han2015deep}
Song Han, Huizi Mao, and William~J. Dally.
\newblock Deep compression: Compressing deep neural networks with pruning,
  trained quantization and huffman coding.
\newblock 2016.

\bibitem{jain2018compensated}
S.~{Jain}, S.~{Venkataramani}, V.~{Srinivasan}, J.~{Choi}, P.~{Chuang}, and
  L.~{Chang}.
\newblock Compensated-dnn: Energy efficient low-precision deep neural networks
  by compensating quantization errors.
\newblock In {\em 55th ACM/ESDA/IEEE Design Automation Conference (DAC)}, pages
  1--6, 2018.

\bibitem{VOS_DSP}
Y.~{Liu}, T.~{Zhang}, and K.~K. {Parhi}.
\newblock Computation error analysis in digital signal processing systems with
  overscaled supply voltage.
\newblock {\em IEEE Transactions on Very Large Scale Integration (VLSI)
  Systems}, 18(4):517--526, 2010.

\bibitem{Salsa}
S.~{Venkataramani}, A.~{Sabne}, V.~{Kozhikkottu}, K.~{Roy}, and
  A.~{Raghunathan}.
\newblock Salsa: Systematic logic synthesis of approximate circuits.
\newblock In {\em DAC Design Automation Conference 2012}, pages 796--801, 2012.

\bibitem{EvolApproach}
Z.~{Vasicek} and L.~{Sekanina}.
\newblock Evolutionary approach to approximate digital circuits design.
\newblock {\em IEEE Transactions on Evolutionary Computation}, 19(3):432--444,
  2015.

\bibitem{MACACO}
R.~{Venkatesan}, A.~{Agarwal}, K.~{Roy}, and A.~{Raghunathan}.
\newblock Macaco: Modeling and analysis of circuits for approximate computing.
\newblock In {\em 2011 IEEE/ACM International Conference on Computer-Aided
  Design (ICCAD)}, pages 667--673, 2011.

\bibitem{SpecAdd}
Ajay~K. Verma, Philip Brisk, and Paolo Ienne.
\newblock Variable latency speculative addition: A new paradigm for arithmetic
  circuit design.
\newblock In {\em Proceedings of the Conference on Design, Automation and Test
  in Europe}, page 1250–1255, 2008.

\bibitem{Truncation-ErrTolerant}
Ning Zhu, Wang~Ling Goh, Weija Zhang, Kiat~Seng Yeo, and Zhi~Hui Kong.
\newblock Design of low-power high-speed truncation-error-tolerant adder and
  its application in digital signal processing.
\newblock {\em IEEE Transactions on Very Large Scale Integration (VLSI)
  Systems}, 18(8), 2010.

\bibitem{IMPACT}
V.~{Gupta}, D.~{Mohapatra}, S.~P. {Park}, A.~{Raghunathan}, and K.~{Roy}.
\newblock Impact: Imprecise adders for low-power approximate computing.
\newblock In {\em IEEE/ACM International Symposium on Low Power Electronics and
  Design}, pages 409--414, 2011.

\bibitem{LogicCompr}
I.~{Qiqieh}, R.~{Shafik}, G.~{Tarawneh}, D.~{Sokolov}, and A.~{Yakovlev}.
\newblock Energy-efficient approximate multiplier design using bit
  significance-driven logic compression.
\newblock In {\em Design, Automation Test in Europe Conference Exhibition
  (DATE), 2017}, pages 7--12, 2017.

\bibitem{Kulkarni}
P.~{Kulkarni}, P.~{Gupta}, and M.~{Ercegovac}.
\newblock Trading accuracy for power with an underdesigned multiplier
  architecture.
\newblock In {\em 2011 24th Internatioal Conference on VLSI Design}, pages
  346--351, 2011.

\bibitem{InAccCounter_Lin}
C.~{Lin} and I.~{Lin}.
\newblock High accuracy approximate multiplier with error correction.
\newblock In {\em 2013 IEEE 31st International Conference on Computer Design
  (ICCD)}, pages 33--38, 2013.

\bibitem{LogApprox}
Patricio~Bulić Uroš~Lotrič.
\newblock Applicability of approximate multipliers in hardware neural networks.
\newblock {\em Neurocomputing}, 96:57--65, 2012.

\bibitem{AlphabetSet}
Syed~Shakib Sarwar, Swagath Venkataramani, Aayush Ankit, Anand Raghunathan, and
  Kaushik Roy.
\newblock Energy-efficient neural computing with approximate multipliers.
\newblock 14, 2018.

\bibitem{AlphSet_DSP}
{Jongsun Park}, {Hunsoo Choo}, K.~{Muhammad}, {SeungHoon Choi}, {Yonghee Im},
  and {Kaushik Roy}.
\newblock Non-adaptive and adaptive filter implementation based on sharing
  multiplication.
\newblock In {\em 2000 IEEE International Conference on Acoustics, Speech, and
  Signal Processing. Proceedings (Cat. No.00CH37100)}, volume~1, pages 460--463
  vol.1, 2000.

\bibitem{Yeo}
{Khaing Yin Kyaw}, {Wang Ling Goh}, and {Kiat Seng Yeo}.
\newblock Low-power high-speed multiplier for error-tolerant application.
\newblock In {\em 2010 IEEE International Conference of Electron Devices and
  Solid-State Circuits (EDSSC)}, pages 1--4, 2010.

\bibitem{TrucSwartzlander}
M.~J. {Schulte} and E.~E. {Swartzlander}.
\newblock Truncated multiplication with correction constant [for dsp].
\newblock In {\em Proceedings of IEEE Workshop on VLSI Signal Processing},
  pages 388--396, 1993.

\bibitem{AreaEffGuibaly}
S.~S. {Kidambi}, F.~{El-Guibaly}, and A.~{Antoniou}.
\newblock Area-efficient multipliers for digital signal processing
  applications.
\newblock {\em IEEE Transactions on Circuits and Systems II: Analog and Digital
  Signal Processing}, 43(2):90--95, 1996.

\bibitem{FixWidthChen}
{Jer Min Jou}, {Shiann Rong Kuang}, and {Ren Der Chen}.
\newblock Design of low-error fixed-width multipliers for dsp applications.
\newblock {\em IEEE Transactions on Circuits and Systems II: Analog and Digital
  Signal Processing}, 46(6):836--842, 1999.

\bibitem{Narayanmoorthy}
S.~{Narayanamoorthy}, H.~A. {Moghaddam}, Z.~{Liu}, T.~{Park}, and N.~S. {Kim}.
\newblock Energy-efficient approximate multiplication for digital signal
  processing and classification applications.
\newblock {\em IEEE Transactions on Very Large Scale Integration (VLSI)
  Systems}, 23(6):1180--1184, 2015.

\bibitem{DRUM}
S.~{Hashemi}, R.~I. {Bahar}, and S.~{Reda}.
\newblock Drum: A dynamic range unbiased multiplier for approximate
  applications.
\newblock In {\em 2015 IEEE/ACM International Conference on Computer-Aided
  Design (ICCAD)}, pages 418--425, 2015.

\bibitem{PP_Perf}
G.~{Zervakis}, K.~{Tsoumanis}, S.~{Xydis}, D.~{Soudris}, and K.~{Pekmestzi}.
\newblock Design-efficient approximate multiplication circuits through partial
  product perforation.
\newblock {\em IEEE Transactions on Very Large Scale Integration (VLSI)
  Systems}, 24(10):3105--3117, 2016.

\bibitem{RedPrec_FP}
J.~Y.~F. {Tong}, D.~{Nagle}, and R.~A. {Rutenbar}.
\newblock Reducing power by optimizing the necessary precision/range of
  floating-point arithmetic.
\newblock {\em IEEE Transactions on Very Large Scale Integration (VLSI)
  Systems}, 8(3):273--286, 2000.

\end{thebibliography}
